\documentclass[acmsmall,screen]{acmart} 

\AtBeginDocument{%
  \providecommand\BibTeX{{%
    \normalfont B\kern-0.5em{\scshape i\kern-0.25em b}\kern-0.8em\TeX}}}

\setcopyright{acmcopyright}
\copyrightyear{2022}
\acmYear{2022}
\acmDOI{XXXXXXX}

\acmJournal{JACM}
\acmVolume{37}
\acmNumber{4}
\acmArticle{111}
\acmMonth{8}

\usepackage{booktabs} 
\usepackage[noend]{algorithm2e}
\usepackage{xcolor}
\usepackage{lipsum}  
\usepackage{subfigure}
\newcommand{\pparam}[0]{\phi}
\newcommand{\cparam}[0]{\theta}

\newcommand{\longname}[0]{Policy Gradient Assisted MAP-Elites}
\newcommand{\sourcecode}[0]{\url{https://github.com/adaptive-intelligent-robotics/pga-map-elites}}
\newcommand{\name}[0]{PGA-MAP-Elites}
\newcommand{\nevo}[0]{p_{evo}}
\usepackage{enumitem}
\usepackage{multirow}

\SetCommentSty{mycommfont}

\begin{document}

\title{Empirical analysis of \name{} for Neuroevolution in Uncertain Domains}

\author{Manon Flageat}
\email{manon.flageat18@imperial.ac.uk}

\author{Felix Chalumeau}
\email{felix.chalumeau20@imperial.ac.uk}

\author{Antoine Cully}
\email{a.cully@imperial.ac.uk}
\affiliation{
  \institution{Imperial College London}
  \department{Department of Computing, Adaptive and Intelligent Robotics Lab}
  \city{London} 
  \country{UK} 
  \postcode{SW7 2AZ}
}


\begin{abstract}
Quality-Diversity algorithms, among which MAP-Elites, have emerged as powerful alternatives to performance-only optimisation approaches as they enable generating collections of diverse and high-performing solutions to an optimisation problem. However, they are often limited to low-dimensional search spaces and deterministic environments.
The recently introduced \longname{} (\name{}) algorithm overcomes this limitation by pairing the traditional Genetic operator of MAP-Elites with a gradient-based operator inspired by Deep Reinforcement Learning.
This new operator guides mutations toward high-performing solutions using policy-gradients. 
In this work, we propose an in-depth study of \name{}. We demonstrate the benefits of policy-gradients on the performance of the algorithm and the reproducibility of the generated solutions when considering uncertain domains.
We first prove that \name{} is highly performant in both deterministic and uncertain high-dimensional environments, decorrelating the two challenges it tackles.
Secondly, we show that in addition to outperforming all the considered baselines, the collections of solutions generated by \name{} are highly reproducible in uncertain environments, approaching the reproducibility of solutions found by Quality-Diversity approaches built specifically for uncertain applications. 
Finally, we propose an ablation and in-depth analysis of the dynamic of the policy-gradients-based variation. We demonstrate that the policy-gradient variation operator is determinant to guarantee the performance of \name{} but is only essential during the early stage of the process, where it finds high-performing regions of the search space.
\end{abstract}

\ccsdesc[500]{Computing methodologies~Evolutionary robotics}

\keywords{Quality-Diversity, MAP-Elites, Reinforcement Learning, Neuroevolution}

\maketitle

\section{Introduction}

Natural evolution can generate diverse species with distinct characteristics existing within the same ecological niche. This variety of survival strategies adopted by cohabiting species has allowed life to endure for instance multiple mass extinction events.
Inspired by this importance of diversity, researchers have proposed Quality-Diversity (QD) optimisation \cite{Cully2018, chatzilygeroudis2021quality, Pugh2016}. 
This branch of Evolutionary Computation does not generate one solution to a problem based on performance only, but rather a collection of diverse and high-performing solutions. 
QD approaches, and in particular, the Multi-dimensional Archive of Phenotypic Elites (MAP-Elites) algorithm \cite{Mouret2015}, have shown promising results in a wide range of domains, from content generation for video games \cite{gravina_procedural_2019}, and aerodynamic design \cite{gaier2017aerodynamic} to robotics \cite{Cully2015}. They allow the discovery of creative solutions \cite{gaier2017aerodynamic, gravina_procedural_2019}, but also non-trivial stepping stones toward even better solutions \cite{Cully2018}, and enable fast adaptation to damages or unseen situations \cite{Cully2015, chatzilygeroudis2021quality}.

QD algorithms traditionally rely on elitism. This is beneficial in environments where performance scores are deterministically associated with solutions; in other words, environments in which a given solution always exhibits the same behaviour. 
However, many environments, refered to as uncertain, do not fulfil this assumption: real robots, for example, use imperfect sensors and actuators. In such environments, a given solution might exhibit slightly different behaviours from one evaluation to another with different performance scores. 
As a consequence, solutions might be lucky and get a higher performance score than they could expect on average. The elitism of QD algorithms preserves these lucky solutions, making uncertain applications an important challenge for QD \cite{Flageat_2020, justesen2019map}. 
In particular, some solutions might be more robust than others to the source of uncertainty and thus be more reproducible. For example, if one actuator is an important source of uncertainty, solutions relying on this actuator would be less reproducible than others. Most QD algorithms do not have mechanism to favour reproducible solutions \cite{Flageat_2020, justesen2019map}.
Besides, QD algorithms are traditionally driven in their exploration of the search space by a Genetic Algorithm (GA). Due to its sample inefficiency, this mechanism makes it difficult -if not impossible- to apply QD to high-dimensional problems \cite{such2017deep}, making it another important limitation. 
This paper focus on domains that are both high-dimensional and uncertain.

On the other side, Deep Reinforcement Learning (DRL) approaches \cite{arulkumaran2017deep} show encouraging results in a wide range of uncertain domains with high-dimensional search-spaces \cite{Mnih2015, Lillicrap2016}.
Unlike QD, DRL aims to find a single performance-maximising solution. It relies on the powerful directed-search abilities of gradient-based learning, as well as the function-approximation capabilities of Deep Neural Networks (DNNs) to find a high-performing and highly-reproducible solution \cite{Lillicrap2016, Fujimoto2018}.
\longname{} (\name{}) \cite{nilsson2021policy} has been recently introduced as an alternative algorithm mixing QD and DRL to scale QD algorithms to large neuroevolution problems. 
\name{} extends MAP-Elites with a new variation operator based on policy-gradients (PG). Half of the solutions are generated using this PG variation operator and the other half using the usual GA variation operator. 

This work proposes an in-depth study of \name{} and, more generally, the benefits of integrating PG in MAP-Elites. We focus on the same tasks as in the \name{} paper, referred to as QD-Gym \cite{nilsson2021policy}. These tasks are all uncertain locomotion tasks with high-dimensional parameter spaces. 
Our contributions can be summarised as followed:
\begin{itemize}
	\item A comparison of \name{} to a larger set of QD and DRL baselines, demonstrating extensively that it maintains the divergent search capabilities of MAP-Elites while finding solutions as performing as those found by compared DRL approach. 
	\item Additional results in deterministic QD-Gym environments, highlighting the benefit of \name{} applied to high-dimensional search-space independently of the uncertainty.
	\item A reproducibility-study in uncertain environment, demonstrating that \name{} does not only generate solutions that are better performing but also more reproducible in uncertain environments, matching and sometimes out-performing the performance of QD algorithms specifically designed for such applications.
	\item To support these analysis, we also propose a new set of loss metrics quantifying the reproducibility of solutions found by QD algorithms in uncertain domains.
	\item An ablation of the proportion of PG and GA variation in \name{}, proving the importance of both type of variations.
	\item An analysis of the impact of PG and GA variation operators on improvement of the collection showing that PG is only essential during the early stage of the process, where it finds high-performing regions of the search space.
\end{itemize}

\begin{figure}[t!]
\centering  
\includegraphics[width = 0.98\hsize]{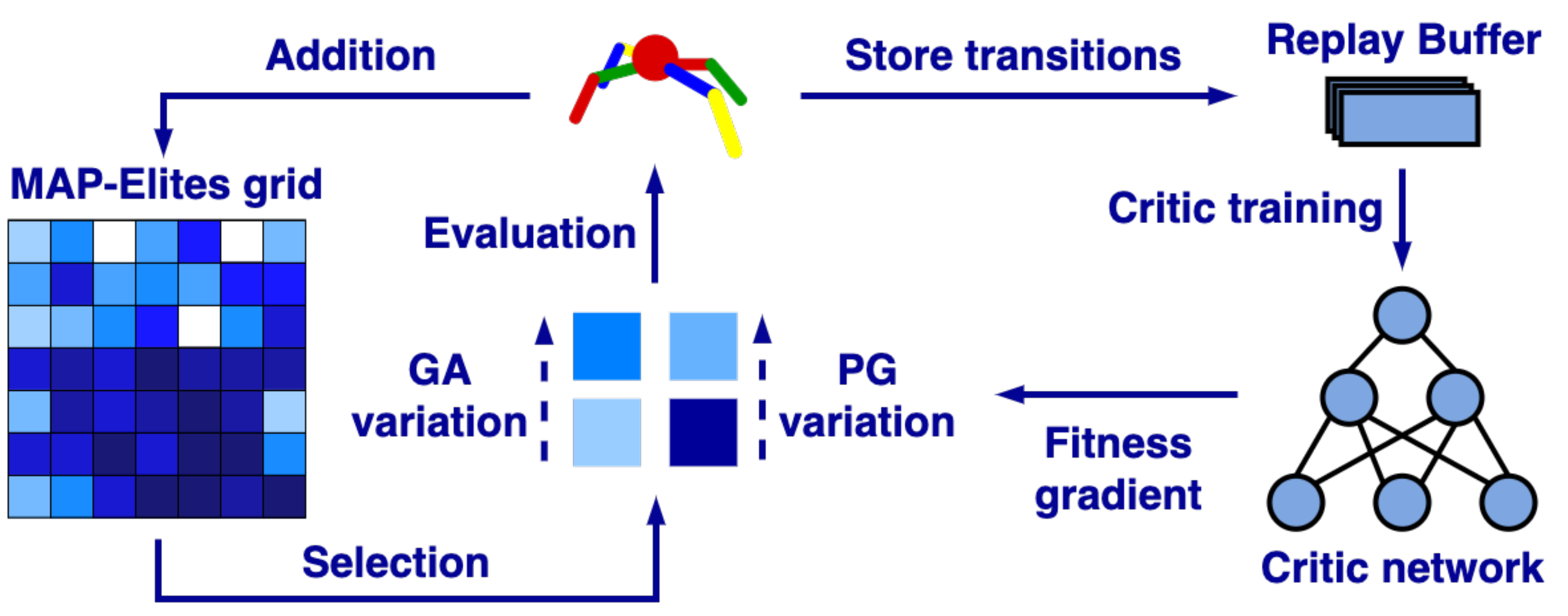}
\caption{
\name{} is based on the standard MAP-Elites loops: parent solutions are selected from the archive and mutated to generate offspring that are evaluated in the environment and added back to the archive. However, in \name{}, the mutation relies on two distinct variation operators: (1) a Policy Gradient (PG) operator that directs variation toward high-performing solutions, and (2) a Genetic Algorithm (GA) operator that maintains divergent search. The PG operator relies on a critic neural network, trained asynchronously to the MAP-Elites loop using experience collected in a replay buffer during evaluations.
}
\label{fig:\name_graphic}
\end{figure}



\section{Background and Related Work}

\subsection{Quality-Diversity Optimisation and MAP-Elites algorithm} \label{sec:map_elites}

While standard optimisation searches for a single high-performing solution, QD optimisation allows finding a collection of solutions that are as high-performing and diverse as possible.
To this end, QD algorithms distinguish solutions that produce different behaviours.
This distinction is based on metrics chosen as part of the task definition to be meaningful for the type of diversity sought. Each metric corresponds to a dimension of interest, denoted as Behavioural Descriptor (BD) \cite{Cully2018, chatzilygeroudis2021quality, Pugh2016}.
For example, different solutions to a robotic locomotion task might induce different gaits. To characterise each gait, the BD of a solution could be the proportion of time each foot of the robot is in contact with the ground \cite{Cully2015, VassiliadesCM16, Colas2020}.
QD algorithms attempt to find solutions as distant as possible in terms of BD while being high-performing in their local region of the BD space; in other words, collections of both diverse and high-performing solutions.
There are currently two leading QD approaches: Novelty Search with Local Competition \cite{Lehman2011} and MAP-Elites~\cite{Mouret2015}, the focus of this work.

The core idea of MAP-Elites is to discretise the BD-space into a grid.
This grid constitutes the final collection of solutions returned at the end of the algorithm. Each of its cells corresponds to a behavioural niche and can store one solution: the elite of this niche.
MAP-Elites aims to fill as many niches as possible with the highest-performing solutions possible. 
The grid is created empty and initially filled with a set of randomly-generated solutions.
One iteration of MAP-Elites can be summarised as: (1) randomly selecting parent solutions from the grid, (2) applying variations (such as mutations or crossovers) to generate offspring, and evaluating them, (3) adding to the grid offspring that either populate an empty cell or outperform an existing elite.

\subsection{Deep Reinforcement Learning setting and TD3 algorithm} \label{sec:td3}

\subsubsection{Reinforcement Learning (RL)}

In the typical RL setting, an agent strives to find the optimal way of acting in an environment.
The agent-environment interaction is modelled as a Markov Decision Process (MDP) given as $<\mathcal{S}, \mathcal{A}, \mathcal{P}, r, \gamma>$. 
The agent acts sequentially at discrete time-steps $t \in [0, T]$, performing actions $\mathbf{a_t} \in \mathcal{A}$. These actions induce transitions detailed by $\mathcal{P}$ between states $\mathbf{s_t}$ and $\mathbf{s_{t+1}} \in \mathcal{S}$ of the environment, and each transition generates a reward $r_{t}$. 
In an MDP, the immediate next state $\mathbf{s_{t+1}}$ only depends on the current state $\mathbf{s_t}$ of the environment.
Internally, the agent chooses its next action based on the current state $\mathbf{s_t}$ following a policy $\pi(\mathbf{a_{t}} | \mathbf{s_{t}})$. The agent aims to learn the policy that  maximises the expected return $\mathbb{E}_{\pi_{\mathbf{\pparam}}}\left[\sum_{t=0}^{T} \gamma^{t} r_{t} \right]$,  where $\gamma \in \left[0,1 \right]$ is the discount factor which regulates importance of future rewards.

\subsubsection{Deep Reinforcement Learning (DRL)}

DRL solves continuous and large-dimensional RL applications by leveraging the powerful directed-search abilities of gradient-based learning, as well as the function-approximation capabilities of Deep Neural Network (DNNs).
Most DRL algorithm rely on approximating with a DNN the action-value function $Q^{\pi_{\mathbf{\pparam}}}(\mathbf{s_t}, \mathbf{a_t})=\mathbb{E}_{\pi_{\mathbf{\pparam}}}\left[\sum_{k=0}^{T-t} \gamma^{k} r_{t+k+1} \mid \mathbf{s_t}, \mathbf{a_t}~\right]$, which encodes the expected return from following the policy $\pi_{\mathbf{\pparam}}$ after having performed action $\mathbf{a_t}$ in state $\mathbf{s_t}$. 
The DNN is trained to approximate $Q^{\pi_{\mathbf{\pparam}}}$ via the Bellman equation in Eq.~\ref{eq:bellman}: it incrementally learns a better approximation of $Q^{\pi_{\mathbf{\pparam}}}$ by bootstrapping from its current estimates.
\begin{equation}
    Q^{\pi_{\pparam}}(\mathbf{s_t}, \mathbf{a_t})=r\left(\mathbf{s_t}, \mathbf{a_t}\right) + \gamma \mathbb{E}\left[Q^{\pi_{\mathbf{\pparam}}}\left(\mathbf{s_{t+1}}, \pi_{\mathbf{\pparam}}\left(\mathbf{s_{t+1}}\right)\right)\right]
\label{eq:bellman}
\end{equation}
In most DRL algorithms, the policy $\pi_{\mathbf{\pparam}}$ is also modelled by a DNN.
The DNN approximating $Q^{\pi_{\mathbf{\pparam}}}$ is often referred to as the critic while the DNN approximating $\pi_{\mathbf{\pparam}}$ as the actor. Methods using both an actor network and a critic network are thus known as actor-critic.

\subsubsection{Twin Delayed Deep Deterministic policy-gradient algorithm (TD3)}

This work is based on TD3 \cite{Fujimoto2018}, one of the state-of-the-art actor-critic DRL algorithms. 
In TD3, the policy $\pi_{\mathbf{\pparam}}$ is learned by an actor network, while the approximation of $Q^{\pi_{\mathbf{\pparam}}}$ is done simultaneously by two distinct critics: $Q_{\mathbf{\cparam_1}}$ and $Q_{\mathbf{\cparam_2}}$. These two critics have the same structure but are trained separately to counteract overestimations; more information are given below.
In TD3, $Q^{\pi_{\mathbf{\pparam}}}$ is learned in an off-policy fashion, meaning using experience collected by any policy, and not only by the current best policy \cite{10.1007/BF00992699}.
Thus, TD3 stores all encountered transitions $\left(\mathbf{s_t}, \mathbf{a_t}, r\left(\mathbf{s_t}, \mathbf{a_t}\right), \mathbf{s_{t+1}}\right)$ in a replay buffer $\mathcal{B}$ to use them to train the critics.
Based on batch of $N$ transitions sampled from $\mathcal{B}$, the loss $L$ defined in Eq.~\ref{eq:loss} updates the critics towards a target $y$ that approximates Eq.~\ref{eq:bellman}.
\begin{align}
    L\left(\mathbf{\cparam_1}, \mathbf{\cparam_2}\right) & = \left(y-Q_{\mathbf{\cparam_1}}\left(\mathbf{s_{t}}, \mathbf{a_{t}}\right)\right)^2 + \left(y-Q_{\mathbf{\cparam_2}}\left(\mathbf{s_{t}}, \mathbf{a_{t}}\right)\right)^2 \label{eq:loss} \\
    y & = r\left(\mathbf{s_t}, \mathbf{a_t}\right)+\gamma \min _{i=1,2} Q_{\mathbf{\cparam_{i}^{\prime}}}\left(\mathbf{s_{t+1}}, \pi_{\mathbf{\pparam^{\prime}}}\left(\mathbf{s_{t+1}}\right) + \epsilon\right)
    \label{eq:target}
\end{align} 

The target $y$ uses the minimum action-value prediction between the two critics $Q_{\mathbf{\cparam_1}}$ and $Q_{\mathbf{\cparam_2}}$ to counteract overestimations done by the critics and avoid instabilities. This is the reason why there is two critic networks.
On top, TD3 also uses target networks $Q_{\mathbf{\cparam_{1}^{\prime}}}$, $Q_{\mathbf{\cparam_{2}^{\prime}}}$ and $\pi_{\mathbf{\pparam^{\prime}}}$ \cite{Mnih2013, Mnih2015}. Target networks are copies of the original networks, but updated with a delay $\tau$ to amortise variations and provide further stability.
Finally, in the expression of the target $y$, $\mathbf{\epsilon} \sim \operatorname{clip}(\sigma_p\mathcal{N}(0, \mathcal{I}),-c, c)$ is a noise on the policy's action to promote reproducibility in uncertain environments. 

The experiences collected in $\mathcal{B}$ are also used to learn the policy $\pi_{\mathbf{\pparam}}$ to maximise $Q^{\pi_{\mathbf{\pparam}}}$, using an approximation of the deterministic policy-gradient \cite{Silver2014}.
As the policy is deterministic, exploration is achieved by adding Gaussian noise with a variance $\sigma_a$ to the selected action: $\mathbf{a_t} = \pi_{\mathbf{\pparam}}\left(\mathbf{s_{t}} \right) + \sigma_a \mathcal{N}(0, \mathcal{I})$. This exploration noise is applied when interacting with the environment, while the reproducibility noise mentioned earlier is only considered to compute the critics target.

\subsection{Alternative variation operator for Quality Diversity} \label{sec:high_dimensional}

In MAP-Elites, as in standard GAs, variations are fitness-agnostic and aim to explore around existing solutions. 
However, this lack of directed search can cause slow convergence even in low-dimensional search space \cite{CMA_ME}, and this problem is increasingly prominent as the number of dimensions increases \cite{Colas2020}. 
Previous works in QD have proposed alternatives to the standard GA mechanism for MAP-Elites. 
\textbf{Covariance Matrix Adaptation MAP-Elites (CMA-MAP-Elites)} \cite{CMA_ME} proposes a directed variation scheme relying on Covariance Matrix Adaptation Evolution Strategy \cite{CMAES}. This approach allows directing the variations toward regions of the search space that maximise the archive improvements and has proven to outperform MAP-Elites in multiple domains~\cite{CMA_ME,cully2020multiemitter}. However, it relies on covariance matrix inversions, which do not scale easily to high-dimensional space.
Alternatively, \textbf{Policy Manifold Search} \cite{rakicevic2021policy} and \textbf{Data-Driven Encoding} \cite{gaier2020discovering} proposed mutations scheme relying on learned low-dimensional representation of the solutions space to scale QD to high-dimensional search-space. 
Another approach, \textbf{MAP-Elites with Evolution Strategies (MAP-Elites-ES)} \cite{Colas2020} replaces GA with Evolution Strategies proposed in \citet{Salimans2017} to extend MAP-Elites to high-dimensional search spaces. This method directs the variation toward an objective using empirical gradients estimated from evaluating a range of perturbations around a current solution. MAP-Elites-ES proposes to use a mixture of fitness and diversity objectives to direct the search. 
However, it requires a large number of evaluations and typically finds archives containing far fewer behaviours than MAP-Elites given an equal amount of experience in the environment \cite{Colas2020}.
Recently, Fontaine et al. have proposed the \textbf{Differentiable-QD} framework \cite{fontaine2021differentiable} for the specific case of QD applied to differentiable domains. This work augments QD search with the information given by the explicit gradient of the function to optimise.
However, it is limited to differentiable domains, which is not the case with DRL, thus it is not directly applicable to the domains considered in this work.
Differentiable-QD has recently been extended to non-differentiable domain in \citet{tjanaka2022approximating} but the authors prove that this approach does not work well for the domains considered in this work.
Finally, \textbf{Quality-Diversity Policy-Gradient (QD-PG)} \cite{pierrot2022diversity} has been recently introduced as a way to combine QD and DRL. 
QD-PG introduces a policy-gradient for diversity, computed using BDs defined at the state level and referred to as state-descriptors. 
At each generation of QD-PG, half of the solutions are updated using this diversity policy-gradient and the other half using the quality policy-gradient.

\subsection{Quality-Diversity applied to uncertain environments} \label{sec:uncertain_env}

In many environments, the exact same solution might not exhibit the same behaviour when evaluated twice. One solution might thus get different fitness or BD values from one evaluation to another.
Such environments have been extensively studied in the Evolutionary Algorithm literature \cite{ea_uncertain}, we refer to them as uncertain environments.

\subsubsection{Performance estimation in uncertain environments}

The lack of reproducibility in uncertain environments directly impacts the performance of QD algorithms. 
In MAP-Elites, solutions are evaluated once and the result of this unique evaluation is used to determine their cell and their fitness. However, this single evaluation might be "lucky" and display higher fitness or higher novelty than it would display on average if more evaluations were performed.
Due to this uncertainty, solutions might be kept in the wrong cell, leading to a loss in the actual coverage of the BD-space.
Lucky solutions might also be kept in the place of truly good-performing ones, resulting in a loss in the actual quality of the collection.
To handle such uncertain environments, the most widely used approach is to sample multiple times solutions \cite{cully2018hierarchical}. In this approach, that we refer to as \textbf{MAP-Elites-sampling}, each solution is evaluated a fixed number of times $M$ before being added to the archive. The fitness and BD of a solution are approximated using the mean fitness and BD of these re-evaluations. However, this approach is costly and impact drastically the algorithm speed-of-convergence. 
Thus, \citet{justesen2019map} proposed \textbf{Adaptive-sampling}: an alternative sampling-based approach that distributes samples more wisely across solutions. This solution mitigates the cost of MAP-Elites-sampling but stays significantly slower in convergence than the original MAP-Elites.
Alternatively, \citet{Flageat_2020} proposed \textbf{Deep-grid}, a MAP-Elites variant that tackle uncertain environments without sampling. 
In Deep-grid, sampling is done implicitly by storing $D$ previously encountered solutions in each cell and considering them as samples of the same solution. 

\subsubsection{Reproducibility in uncertain environments}

Most uncertainty-handling approaches focus on computing a better approximation of the "ground-truth" fitness and BD of solutions. However, in certain tasks, some solutions are more reproducible than others and thus more robust to uncertainty. Said differently, while it is important to approximate the expected value of the fitness and BD distributions of each solution, it is as valuable to minimise the variance of these distributions as this leads to more reproducible solutions. A challenge of QD approaches in such domains is to generate solutions as reproducible as possible, while producing large collections of truly diverse and high-performing solutions. 
Adaptive-sampling proposes an explicit mechanism to delete from the grid solutions whose BD is evaluated too often outside of their cell. 
Deep-grid on the other side enforces reproducible solutions via its selection and addition mechanisms. These mechanisms systematically question solutions, even after their addition to the grid, and slowly reject from the grid non-reproducible solutions. 
We hypothesise that algorithms that use empirical estimation of the gradient for variation should enforce reproducible solutions. Gradient approximations are based on sampling and are thus be less sensitive to luck. It is the case of MAP-Elites-ES, described in Section \ref{sec:high_dimensional}, which, on top, also uses $30$ re-sampling of the offspring to tackle uncertainty.
Finally, we demonstrate in this paper that the use of policy-gradient variation operators also promotes reproducible solutions in QD collections. 

\RestyleAlgo{ruled}
\begin{algorithm}[!t]
\caption{\name ~algorithm.}
\label{alg:\name}

\KwData{$n_{gen}$ total generations, $b$ batch-size, $\nevo$ proportion of GA and PG variations, $n_{crit}$ and $n_{act}$ PG parameter, and $\sigma_1$ and $\sigma_2$ GA parameters}
\KwResult{the final archive $\mathcal{A}$}

\For{$i = 0 \rightarrow n_{gen}$}{
    \eIf{$i = 0$}{
        
        $\pi_{\mathbf{\widehat{\pparam}_1}}, \dots, \pi_{\mathbf{\widehat{\pparam}_b}} =\text{random\texttt{\char`_}solutions}()$
    }{
        \tcp{Train critics and greedy actor}
        
        \For{$k = 1  \rightarrow  n_{crit}$}{
            $Q_{\mathbf{\cparam_1}}, Q_{\mathbf{\cparam_2}}, Q_{\mathbf{\cparam_1^{\prime}}}, Q_{\mathbf{\cparam_2^{\prime}}} = \text{train\texttt{\char`_}critics}(\mathcal{B}, Q_{\mathbf{\cparam_1}}, Q_{\mathbf{\cparam_2}}, Q_{\mathbf{\cparam_1^{\prime}}}, Q_{\mathbf{\cparam_2^{\prime}}}, \pi_{\mathbf{\pparam_{greedy}^{\prime}}})$
            
            $\mathbf{\pi_{\pparam_{greedy}}}, \pi_{\mathbf{\pparam_{greedy}^{\prime}}} = \text{train\texttt{\char`_}actor}(\mathcal{B}, Q_{\mathbf{\cparam_1}}, \mathbf{\pi_{\pparam_{greedy}}}, \pi_{\mathbf{\pparam_{greedy}^{\prime}}})$
        }
        \tcp{Apply GA variation}
        \For{$j = 1  \rightarrow \lfloor{\nevo \cdot b}\rfloor$}{
                
            $\pi_{\mathbf{\pparam_{p_1}}}, \pi_{\mathbf{\pparam_{p_2}}}=\text{uniform\texttt{\char`_}selection}(\mathcal{A})$
                
            $\pi_{\mathbf{\widehat{\pparam}_j}} = \pi_{\mathbf{\pparam_{p_1}}}+\sigma_{1}\mathbf{\mathcal{N}}(\mathbf{0},  \mathbf{I})+\sigma_{2}\left(\pi_{\mathbf{\pparam_{p_2}}}-\pi_{\mathbf{\pparam_{p_1}}}\right) \mathcal{N}(0,1)$
        }
        \tcp{Apply PG variation}
        \For{$j = \lfloor{\nevo \cdot b}\rfloor \rightarrow b-1$}{
            $\mathbf{\pi_{\pparam_p}}=\text{uniform\texttt{\char`_}selection}(\mathcal{A})$
                
            \For{$k = 1  \rightarrow  n_{act}$}{
                
                $\pi_{\mathbf{\widehat{\pparam_j}}} = \text{train\texttt{\char`_}actor}(\pi_{\mathbf{\widehat{\pparam_j}}}, \mathcal{B}, Q_{\mathbf{\cparam_1}}, \mathbf{\pi_{\pparam_p}})$
                    
            }
        }
        \tcp{Return the greedy actor as offspring}
        $\mathbf{\widehat{\pparam}_b} = \mathbf{\mathbf{\pparam}_{greedy}}$
	}
	
	\tcp{Evaluate offspring and update $\mathcal{B}$ and $\mathcal{A}$}
	
	$fit, bd, transitions = \text{evaluate}(\pi_{\widehat{\pparam}_1}, ... \pi_{\widehat{\pparam}_b})$ 
	
	$\mathcal{B} = \text{add\texttt{\char`_}to\texttt{\char`_}replay\texttt{\char`_}buffer}\left(\mathcal{B}, transitions \right)$
	
	$\mathcal{A} = \text{add\texttt{\char`_}to\texttt{\char`_}archive}\left(\mathcal{A}, fit, bd \right)$

}

\end{algorithm}

\subsection{\name{}}

In this work, we study \name{}, a new QD algorithm proposed by \citet{nilsson2021policy} and illustrated in Fig.~\ref{fig:\name_graphic} and Alg.~\ref{alg:\name}. 
\name{} allows to apply the QD approach to high-dimensional search spaces by combining ideas from MAP-Elites and TD3.
It follows the usual MAP-Elites loop (selection, variation, evaluation, addition) but uses two independent variation operators: a standard Genetic Algorithm (GA) operator from MAP-Elites, and a Policy Gradient (PG) operator inspired by TD3.
The PG operator relies on policy-gradients derived from critics trained in parallel to the main loop. 
In summary, \name{} adds three components to standard MAP-Elites: (1) a replay buffer collecting experience from evaluations, (2) two critic networks and their associated policy, trained using the replay buffer, and (3) a PG variation operator.

\subsubsection{Replay buffer}

At each iteration of the algorithm loop, a fixed number of offspring policies are evaluated in the environment. A replay buffer $\mathcal{B}$ collects every transition $(s_t, a_t, r_t, s_{t+1})$ of these experiences. $\mathcal{B}$ has a limited capacity, and old transitions are removed on a first-in-first-out basis.

\subsubsection{Critics and greedy actor}

Like TD3, \name{} trains two critic networks $Q_{\cparam_1}$ and $Q_{\cparam_2}$ to approximate the action-value function, and an associated "greedy" actor $\pi_{\pparam}$. \name{} keeps $\pi_{\pparam}$ separate from the actors in the archive to compute the critic update target in Eq.~\ref{eq:target}. This way the target is computed using an actor that approximate the optimal action in each state, not conditioned by a bd. 
The two critics and the greedy actor are trained at each MAP-Elites loop for $n_{crit}$ steps of gradient descent, with the loss given by Eq.~\ref{eq:loss} and the same sampled noise $\epsilon$ that implicitly favours reproducible behaviours. 
This part of \name{} is exactly the same as in TD3.
It only starts after initialisation to ensure a good initial distribution of experience in the replay buffer.
Finally, a copy of the greedy controller is always considered as an offspring for evaluation as it may provide useful behaviours.

\subsubsection{PG and GA variation operators}

At each iteration of the algorithm's main loop, the randomly selected parents are updated by variation operators to generate offspring.
The core idea of \name{} is to introduce a new PG variation operator, based on the critic networks $Q_{\cparam_1}$ and $Q_{\cparam_2}$ and the replay buffer $\mathcal{B}$. 
This operator updates a solution by applying $n_{act}$ consecutive steps of gradient ascent based on $N$ transitions sampled uniformly from $\mathcal{B}$. It thus directs variations toward high-performing regions of the search space.
To maintain the divergent search methodology of standard MAP-Elites, the PG variation operator is paired with a GA variation operator, here we use the directional variation introduced by \citet{vassiliades2018discovering}.
This variation is based on two parents sampled uniformly from the archive. It combines a displacement in the parameter space along the line between the two parents and a random perturbation.
The PG and GA operators used in \name{} are independent and used to generate distinct offspring controllers. The proportion of controllers generated by each of them is determined by the parameter $\nevo$.
At each iteration, the next batch of offspring controllers to be evaluated is of size $b$, among which $\lfloor{\nevo \cdot b}\rfloor$ are generated through GA variation and $\lceil(1 - \nevo) \cdot b\rceil$ through PG variation. Among the controllers generated through PG, one is always a copy of the greedy controller associated with the critics.

\section{Experimental methodology}

This work proposes an in-depth analysis of \name{} divided in three main parts:
\begin{itemize}[leftmargin=*]
    \item Section ~\ref{sec:deterministic} studies the performance of \name{} on tasks with high-dimensional search space, and with or without uncertainty on the fitness and BD estimations.
    \item Section ~\ref{sec:stochastic} proposes a study of the reproducibility of the solutions found by \name{}.
    \item Section ~\ref{sec:ablation} displays an ablation of the main parameter of \name{}: the proportion of GA and PG variations; and a study of the dynamics of these variation operators. 
\end{itemize}
All these results use the same environments, baselines, implementations, and metrics. We introduce them in this section.

\subsection{Environments} \label{sec:environments}

\subsubsection{Tasks}
We consider the four QD-Gym tasks defined in \citet{nilsson2021policy}, derived from the standard DRL benchmark OpenAI Gym \cite{Gym} in PyBullet \cite{coumans2019}. In all these tasks, a robot aims to discover all the ways it can walk while maximising a trade-off between speed and energy consumption. 
The details of these tasks are given in Table ~\ref{tab:qd_gym}.

\begin{table}[ht]
  \caption{QDGym tasks used in this work.}
  \label{tab:qd_gym}
  \begin{tabular}{l | c c c c}
    \toprule
    & \textsc{QDWalker} & \textsc{QDHalfCheetah} & \textsc{QDAnt} & \textsc{QDHopper}\\
    & \includegraphics[width = 0.12\textwidth]{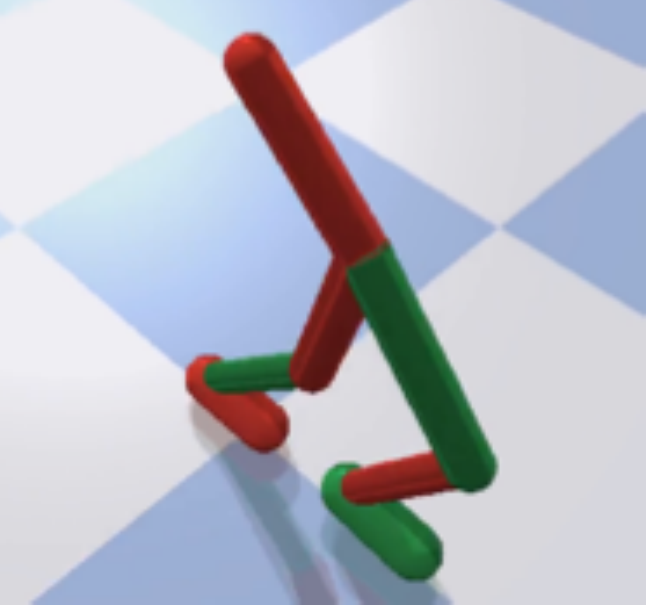} & \includegraphics[width = 0.12\textwidth]{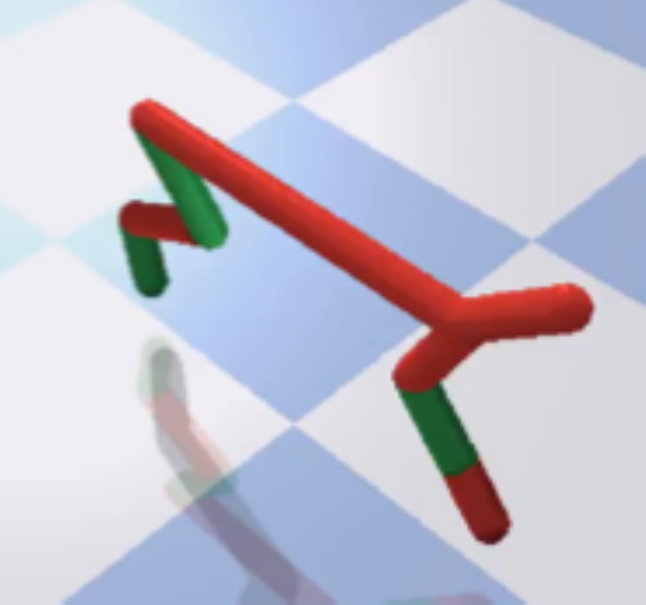} & \includegraphics[width = 0.12\textwidth]{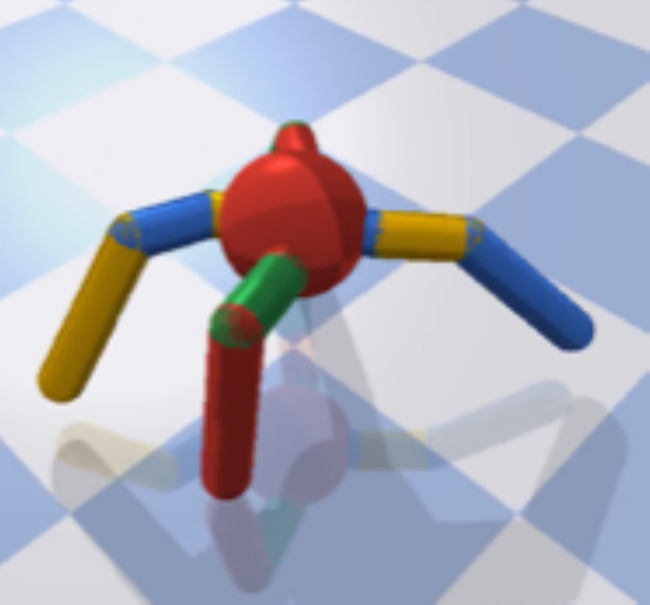} & \includegraphics[width = 0.12\textwidth]{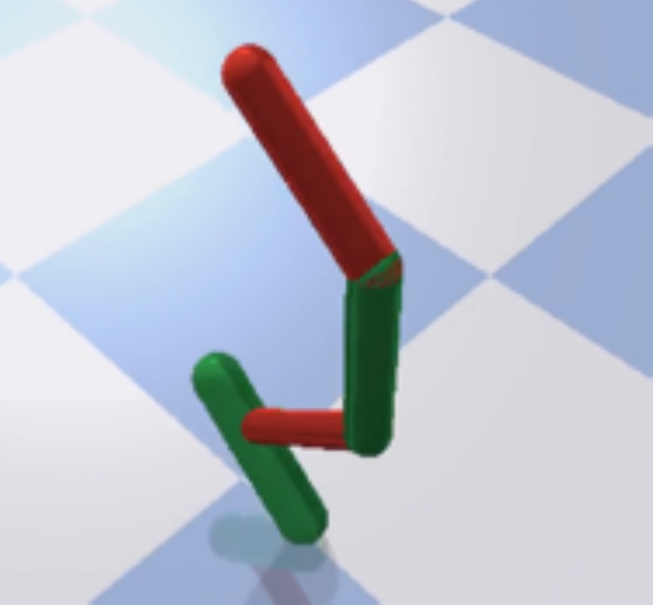} \\
    \midrule
    \textsc{State description} & \multicolumn{4}{c}{Position and velocity of the centre of gravity and joints} \\
    \textsc{Action description} & \multicolumn{4}{c}{Continuous-valued torques in each joint} \\
    \textsc{Fitness definition} & \multicolumn{4}{c}{Forward progress reward + surviving reward + energy usage penalty} \\
    \textsc{BD definition} & \multicolumn{4}{c}{Proportion each foot of the robot is in contact with the ground} \\
    \textsc{State dimensions} & 22 & 26 & 28 & 15 \\
    \textsc{Action dimensions} & 6 & 6 & 8 & 3 \\
    \textsc{BD dimensions} & 2 & 2 & 4 & 1 \\
    \textsc{Nr BD niches} & 1024 & 1024 & 1296 & 1000 \\ 
    \textsc{Episode length} & \multicolumn{4}{c}{1000 simulation steps} \\
    \textsc{Controller} & \multicolumn{4}{c}{Deep Neural Network with 2 hidden layers of 128 neurons} \\ 
    \textsc{Nr parameters} & 20230 & 20742 & 21256 & 18947 \\
  \bottomrule
\end{tabular}
\end{table}

\subsubsection{Uncertainty}
In the QD-Gym tasks, the initial joint positions are sampled from a Gaussian distribution, making these tasks uncertain.
This stochasticity has a limited impact and intuitively one might think that controllers should reach a stationary gait, independent of the initialisation. 
However, these stationary behaviours can only be reached at a long horizon, and the simulation time is limited to a few seconds. In addition, for some controllers, the initialisation can lead to a fall which simply stops the simulation. 
Thus the fitness and BD of a given controller are computed based on a transitional gait, and as such, they are uncertain (see Section \ref{sec:uncertain_env}).
In addition, we hypothesise that, in the QD-Gym tasks, some controllers are more reproducible than others and thus more robust to this uncertainty.

\subsection{Baselines} \label{sec:baselines}

In the next sections, we consider baselines among the list below, also detailed in Table \ref{tab:baselines}.
All the baselines we re-implemented use a CVT-shaped MAP-Elites-archive \cite{VassiliadesCM16} to store solutions. For MAP-Elites-ES and QD-PG only we use the original implementation from the authors with standard MAP-Elites grid-based archive.
All baselines use the same number of cells so this difference in archive type does not impact the results.
\begin{itemize}
    \item \textbf{MAP-Elites} (Section \ref{sec:map_elites}) with directional variation \cite{vassiliades2018discovering}.
    \item \textbf{TD3} (Section \ref{sec:td3}) the original TD3 algorithm from \cite{Fujimoto2018} augmented with a CVT-MAP-Elites archive, used to passively collect behaviours for comparison purposes.
    \item \textbf{CMA-MAP-Elites} (Section \ref{sec:high_dimensional}) with 5 improvement emitters \cite{fontaine2020covariance}.
    \item \textbf{MAP-Elites-sampling} (Section \ref{sec:uncertain_env}) with $M=10$ samples.
    \item \textbf{Deep-grid} (Section \ref{sec:uncertain_env}) with depth $D=50$ as in the original paper.
    \item \textbf{MAP-Elites-ES} (Section \ref{sec:high_dimensional})
    \item \textbf{QD-PG} (Section \ref{sec:high_dimensional})
\end{itemize} 

\begin{table}[ht]
  \caption{List of the baselines used in this work and which comparison they figure in. We also outline the type of mutation used for each baseline and the origin of the implementation used to generate the results.} 
  \label{tab:baselines}
  \begin{tabular}{l | c c | c c }
    \toprule
    & \textsc{Sec. \ref{sec:deterministic}} 
    & \textsc{Sec.~\ref{sec:stochastic}} 
    & Variation
    & \textsc{Code} \\
    
    
    
    \midrule
    
    \textsc{PGA-MAP-Elites} 
    & \checkmark 
    & \checkmark 
    & GA \& policy-gradient
    & Original code \\
    
    \textsc{MAP-Elites-ES} 
    & \checkmark 
    & \checkmark 
    & natural-gradient
    & Original code \\
    
    \textsc{QD-PG} 
    & \checkmark 
    & 
    & diversity-gradient \& policy-gradient
    & Original code \\
    
    \textsc{CMA-MAP-Elites} 
    &  \checkmark 
    & 
    & CMA-ES
    & PyRibs lib \\
    
    \textsc{TD3} 
    &  \checkmark 
    & 
    & policy-gradient
    & Ours \\
    
    \textsc{MAP-Elites} 
    & \checkmark 
    & \checkmark 
    & GA
    & Ours \\
    
    \textsc{MAP-Elites-sampling} 
    & 
    & \checkmark 
    & GA
    & Ours \\
    
    \textsc{Deep-Grid} 
    & 
    & \checkmark 
    & GA
    & Ours \\
    
  \bottomrule
\end{tabular}
\end{table}

\subsection{Implementation and hyperparameters} \label{sec:hyperparams}

Our source code is available at \sourcecode{}. 
It includes a containerised environment to replicate our experiments.
Our code is based on source code from \citet{nilsson2021policy}.
Common hyperparameters are identical between \name{} and MAP-Elites, TD3, MAP-Elites-sampling and Deep-grid, and correspond to the ones used in the original \name{} paper \cite{nilsson2021policy} apart from two parameters of the PG variation: the number of actor training steps $n_{act}$ and the actor learning rate $lr_{act}$. 
We increased these two parameters values in \name{} only, as it proves to significantly improve its performance.
All hyperparameters for these algorithms are given in Appendix~\ref{app:hyperparam}. 
CMA-MAP-Elites \cite{fontaine2020covariance} is implemented using the PyRibs library proposed by its authors \cite{pyribs} with default parameters.
For MAP-Elites-ES \cite{Colas2020} and QD-PG \cite{pierrot2022diversity} we use the implementation provided by the original authors.

\subsection{Evaluation Metrics} \label{sec:metrics}

In the following sections, we consider the following metrics and report p-values based on the Wilcoxon rank-sum test with Bonferroni correction (summarised in Appendix \ref{app:p-values}).

\subsubsection{Main metrics:} In the following, we consider two main metrics:
\begin{itemize}

    \item \textbf{QD-score:}~\cite{Pugh2016} Sum of all fitnesses of the archive, offsetted by the lowest fitness value. It quantifies the diversity and performance of the final collection and allows comparison with QD approaches.

    \item \textbf{Max-fitness:} Fittest solution in the archive, allow comparison with DRL approaches.
\end{itemize}
For the sake of conciseness, we are not reporting the Coverage metric comparisons within the core of the paper, but we provide them in Appendix \ref{app:coverage}.

\subsubsection{Corrected metrics:} To take into account the effects of uncertainty, we re-evaluate each solution kept by the algorithm a fixed number of times $N$, and use the average fitness and BD of these $N$ re-evaluations as approximation of the “ground truth” fitness and BD \cite{Flageat_2020, justesen2019map}.
We place the re-evaluated solutions in a "Corrected archive", using the same archive addition rules with the “ground truth” fitness and BD. We then compute the QD-score and Max-fitness of this Corrected archive. We refer to them as \textbf{Corrected QD-score} and \textbf{Corrected Max-fitness}.
The $N$ re-evaluations are only used for metric computation: the algorithms cannot access them for optimisation.

\subsubsection{Loss metrics:} We also propose to study the loss induced by the re-evaluations, as a way to estimate the reproducibility of the solutions generated by an algorithm. For each metric, we define the loss as the difference between the original 1-sample-based metric and the Corrected $N$-samples-based metric, normalised by the original value of the metric reached by the algorithm:
\begin{itemize}
    \item \textbf{QD-score loss:} Normalised loss between QD-score and Corrected QD-score.
    \item \textbf{Max-fitness loss:} Normalised loss between Max-fitness and Corrected Max-fitness.
\end{itemize}

\section{Performance of \name{}}
\label{sec:deterministic}

\begin{figure*}[t]
\centering 
\includegraphics[width = 0.97\hsize]{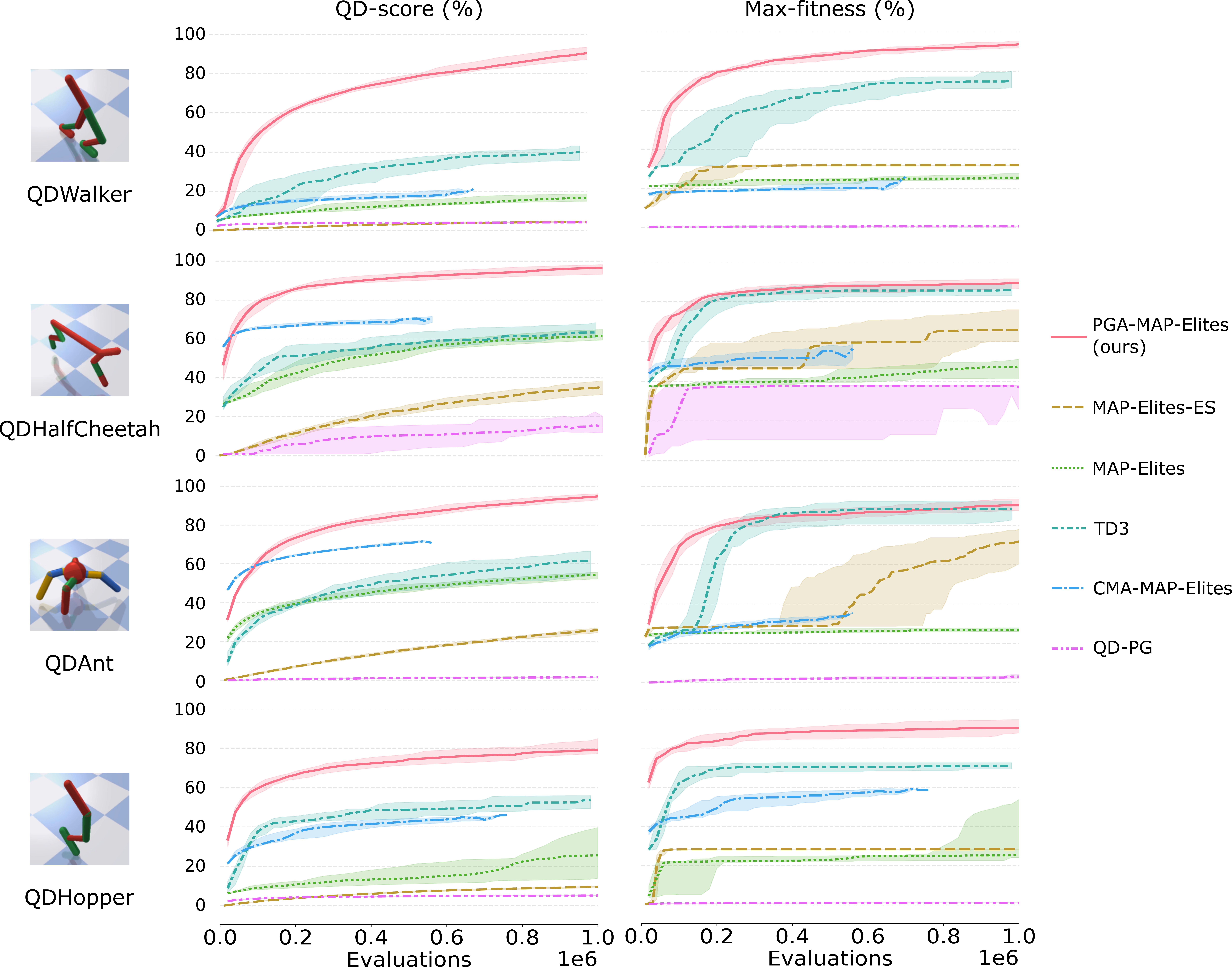}
\caption{
    \textbf{QDGym results:}
    Comparison of the QD-score (left) and Max-fitness (right) of all algorithms on each QDGym task for $10^6$ controller evaluations. Each experiment is replicated $20$ times, the solid line corresponds to the median over replications and the shaded area to the first and third quartiles.
}
\label{fig:progress}
\end{figure*}

We first study the performance of \name{} on the QD-Gym suite, considering the baselines from Section \ref{sec:baselines} and the new hyper-parameters of \name{} described in Section \ref{sec:hyperparams}.
The results are displayed in Fig. \ref{fig:progress}.
The QD-Gym tasks are both high-dimensional and uncertain, so we also propose to decorrelate these two distinct challenges and demonstrate the performance of \name{} on deterministic QD-Gym tasks.
The uncertainty in the QD-Gym tasks arises from the random initialisation of the robots' joints positions. Here, we consider a new version of the QD-Gym tasks where all joints are initialised to $0$ at every episode. The dynamics in PyBullet being deterministic \cite{coumans2019}, this modification leads to a deterministic variant of the QD-Gym environments. 
Fig. \ref{fig:progress_deterministic} displays the performance across all deterministic tasks. Fig. \ref{fig:maps_deterministic} pictures one final archive per algorithm for the Deterministic QDWalker and Deterministic QDHalfCheetah tasks.   
Appendix \ref{app:p-values} summarises the p-values of these comparisons.

Due to computational resource limitations, each run is given a maximum budget of 3 days of computation on 32 cores. All algorithms except CMA-MAP-Elites manage to reach $10^6$ evaluations within this budget. CMA-MAP-Elites relies on costly covariance matrix inversion having $\approx 4.10^8$ parameters, which explains why it does not reach the maximum number of evaluations.

\subsection{QD-score}

\begin{figure*}[t]
\centering
\includegraphics[width = 0.97\hsize]{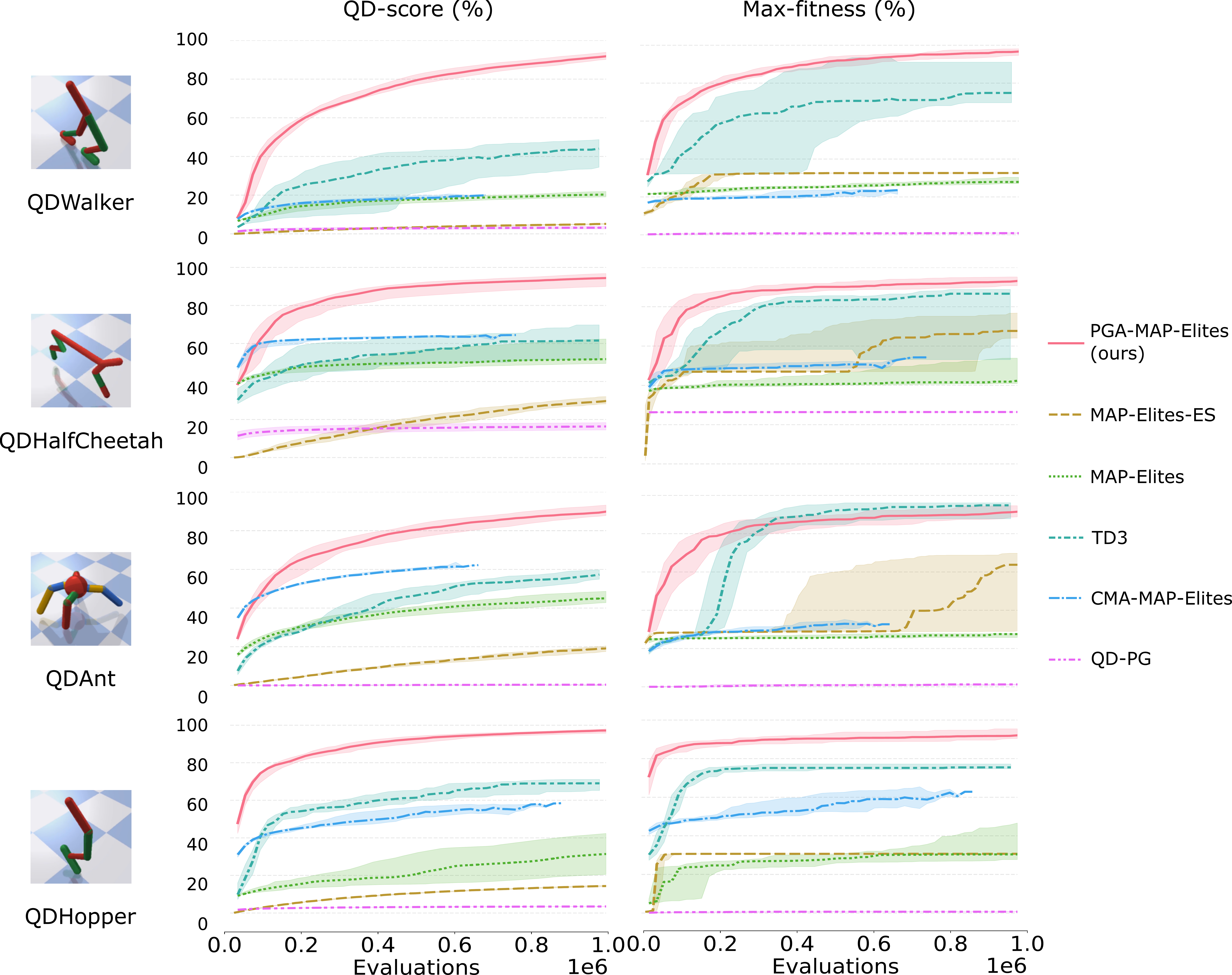}
\caption{
    \textbf{Deterministic QDGym results:}
    Comparison of the QD-score (left) and Max-fitness (right) of all algorithms on each Deterministic task for $10^6$ controller evaluations. Each experiment is replicated $20$ times, the solid line corresponds to the median over replications and the shaded area to the first and third quartiles.
}
\label{fig:progress_deterministic}
\end{figure*}

\begin{figure*}[t]
\centering 
\includegraphics[width = 0.98\hsize]{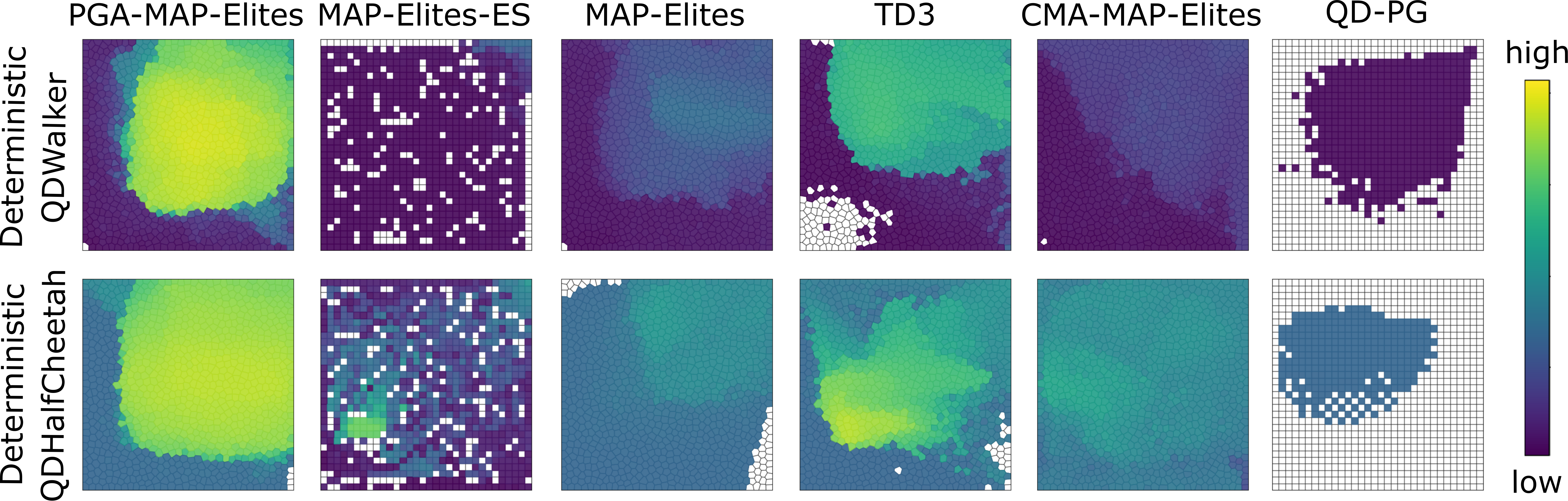}
\caption{
    Final archives in the Deterministic QDWalker (top) and Deterministic QDHalfCheetah (bottom) tasks after $10^6$ controller evaluations. The BD in these tasks are 2-dimensional, giving a square grid. Increasing feet contact time from left to right and from bottom to top. The colour of each cell represents the fitness of the controller it contains (the lighter the better), grey colour corresponds to an empty cell. 
}
\label{fig:maps_deterministic}
\end{figure*}

The QD-score metric in Fig. \ref{fig:progress}  and \ref{fig:progress_deterministic} shows that \name{} outperforms all baselines in terms of quality and diversity of the final archive, and reaches the best collection across all tasks ($p<1\mathrm{e}{-6}$). 
This observation holds on the archives in Fig.~\ref{fig:maps_deterministic}.
MAP-Elites and CMA-MAP-Elites both find significantly less-performing archives than \name{}. 
In the case of CMA-MAP-Elites, this could be due to the early stopping of the algorithm. 
However, in the case of MAP-Elites, it is likely due to the ineffectiveness of GA variation in large search space \cite{Colas2020}.
On the contrary, the PG variation operator of \name{} allows directing the variation toward promising regions of the search space.
Yet, \name{} also outperforms TD3, which also relies on PG. But this result is expected as TD3 only focuses on optimising the best solutions and not the intermediate solutions encountered during the optimisation process that populate the archive. 
\name{} also performs better than QD-PG and MAP-Elites-ES, two QD approaches designed for high-dimensional search-spaces. 
MAP-Elites-ES has a higher sample-cost than \name{} as it relies on sampling to approximate the gradient, and its QD-score does not seem to have fully converged despite the high number of evaluations available. 
On the contrary, QD-PG seems to have converged, but does not succeed to find diverse and high-performing solutions in the considered tasks. 
QD-PG assumes that the sum of the novelty of each state provides a meaningful characterisation for the novelty of the entire behaviour. However, there are no guarantees that it is the case for the QD-Gym tasks.

These results are similar in uncertain and deterministic tasks, except for QDAnt: interestingly, all baselines reach a lower QD-score in the deterministic version of this task  (Fig.~\ref{fig:progress_deterministic}) than in its original version (Fig.~\ref{fig:progress}). 
This task has the larger search space of the QD-Gym domains, and the Ant is the only 4-legged robot in the suite, which might lead to a more complex subspace of promising policies, making it harder to discover and explore.
The randomness of the initial positions might help the algorithms with this exploration, leading to a better QD-Score.
Still, \name{} significantly outperforms all other algorithms in terms of QD-score across all tasks.
According to this first metric, \name{} is the only considered algorithm that finds high-performing solutions throughout the BD-space in all QD-Gym domains.

\subsection{Max-fitness}

\name{} finds the best solutions of all QD approaches ($p<1\mathrm{e}{-6}$).
TD3 also finds these high-performing solutions in QDHalfCheetah and QDAnt and in their deterministic versions, but \name{} outperforms it in QDWalker and QDHopper and their Deterministic versions ($p<1\mathrm{e}{-7}$).
Thus, \name{} retains the optimisation capability of DRL approaches.

\subsection{Conclusion}

These results show that \name{} maintain the capabilities of QD algorithms to generate diverse and high-performing populations while retaining the faculties of DRL approaches to optimise single solutions for performance.
As uncertainty might impact some of the baselines \cite{Flageat_2020}, this section also proves that \name{} outperforms all approaches in Deterministic QD-Gym. Thus, \name{} allows applying the QD-approach to high-dimensional neuroevolution tasks, independently of their uncertain nature.

\section{Reproducibility in uncertain environments} \label{sec:stochastic}

Previous works have shown that QD algorithms applied to uncertain environments tend to favour lucky solutions and struggle to find solutions that are truly diverse and high-performing (see Section \ref{sec:uncertain_env}). On the contrary, the critic networks in TD3 encourage reproducible behaviours (see Section \ref{sec:td3}). As \name{} built on both MAP-Elites and TD3, it should find more reproducible controllers than vanilla QD approaches in uncertain environments.
This section analyses the impact of uncertainty on the performance of \name{}. It compares the reproducibility of the controllers it produces to those of MAP-Elites, MAP-Elites-ES as well as approaches designed specifically for such applications: MAP-Elites-sampling and Deep-grid (see Section \ref{sec:uncertain_env}).

\begin{figure*}[t]
\centering 
\includegraphics[width = 0.98\hsize]{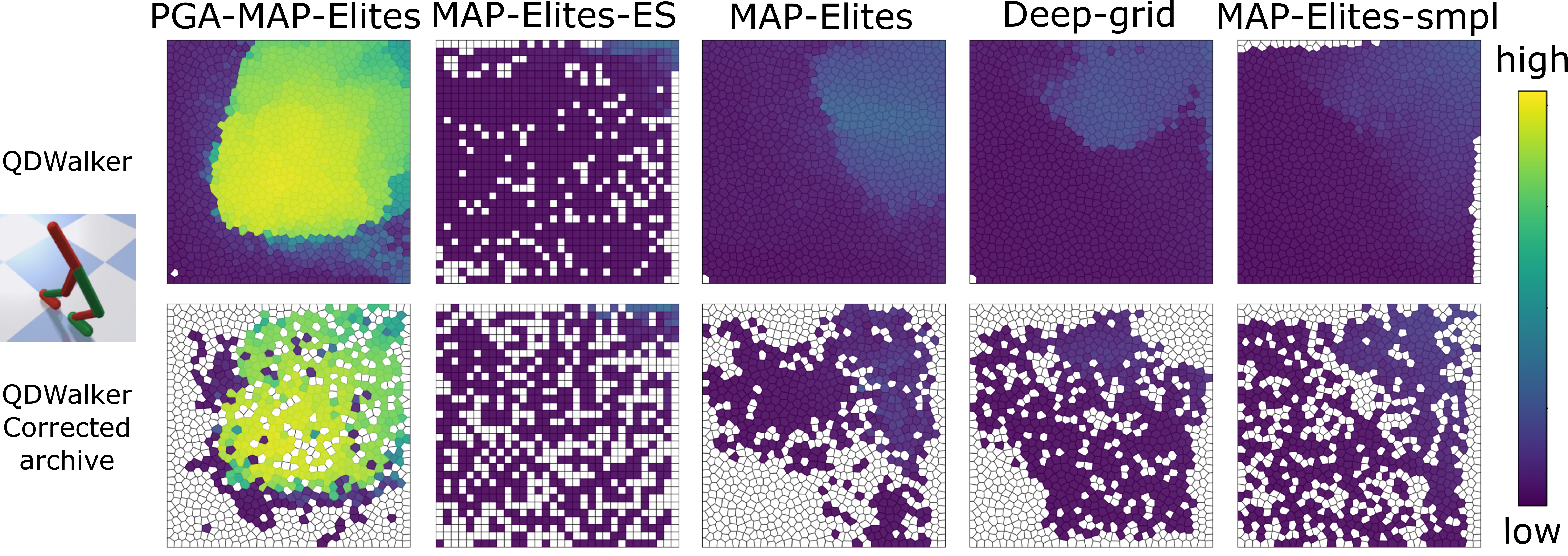}
\caption{
    Final archives (top) and final Corrected archives (bottom) found by all algorithms in the QDWalker task after $10^6$ controller evaluations. The BDs in this task are 2-dimensional, giving a square grid. Increasing feet contact time from left to right and from bottom to top. The colour of each cell represents the fitness of the controller it contains (the lighter the better), grey colour corresponds to an empty cell. 
    The Corrected archive is filled with the controllers from the standard MAP-Elites archive with the same archive-addition rules but using the average fitness and BD over $50$ replications of this controller.
}
\label{fig:maps_stochastic_part}
\end{figure*}

Fig.~\ref{fig:progress_stochastic_part} displays the original and Corrected metrics on the QDWalker task, and Fig.~\ref{fig:maps_stochastic_part} the corresponding original and Corrected archives. As detailed in Section \ref{sec:metrics}, to compute the Corrected metrics, all solutions of the archive are replaced in a Corrected archive according to their average BD and fitness over $50$ replications.
The results for the other tasks of the QD-Gym suite can be found in Appendix~\ref{app:stochastic}.
Fig.~\ref{fig:loss_stochastic} gives the loss in performance induced by uncertainty across all tasks. The loss corresponds to the difference of each Corrected metric in proportion of the original metric. Lower loss indicates better reproducibility.
As already observed in previous uncertainty studies \cite{Flageat_2020, justesen2019map}, the re-evaluation process results in a loss in performance for all the algorithms.

\subsection{Corrected QD-score and QD-score loss}

\begin{figure*}[t!]
\centering
\includegraphics[width = 0.98\hsize]{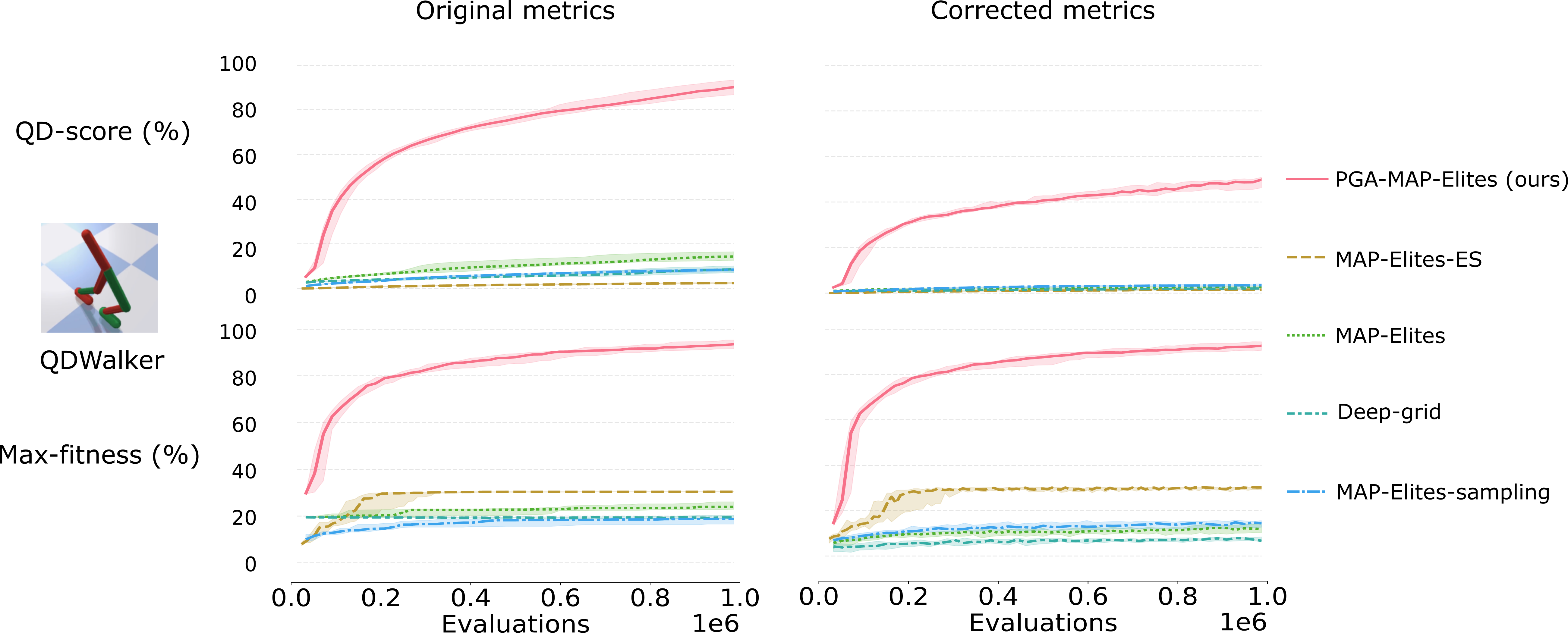}
\caption{
\textbf{Corrected results:}
Comparison of the original (left) and corrected (right) QD-score (top) and Max-fitness (bottom) of all algorithms on QDWalker.
To compute the Corrected metrics, all solutions of the archive are replaced in a Corrected archive according to their average BD and fitness over $50$ replications.
Each experiment is replicated $20$ times, the solid line corresponds to the median over replications and the shaded area to the first and third quartiles. 
}
\label{fig:progress_stochastic_part}
\end{figure*}

The results in Fig.~\ref{fig:progress_stochastic_part} and Appendix~\ref{app:stochastic} show that \name{} outperforms all baselines in Corrected QD-score ($p<1\mathrm{e}{-9}$).
This results is expected as \name{} was already outperforming all other approaches before the re-evaluation in Sec.~\ref{sec:deterministic}.
However, the QD-score loss in Fig.~\ref{fig:loss_stochastic} indicates that \name{} also manages to find reproducible solutions.
In QDWalker, QDHalfCheetah and QDAnt, the QD-score losses of MAP-Elites are consistently higher than those of other algorithms except for Deep-grid ($p<5\mathrm{e}{-7}$).
Only in QDHopper does MAP-Elites observe similar performance-drop. This task seems to produce more reproducible solutions overall.
On the contrary, MAP-Elites-sampling has a significantly lower QD-score loss than MAP-Elites on all tasks ($p<1\mathrm{e}{-7}$). Compared to vanilla MAP-Elites, MAP-Elites-sampling gets a better estimate of the true fitness and BD of a solution through sampling, thus it is less likely to keep lucky solutions which reduce the QD-score-drop. 
The implicit mechanism of Deep-grid fails to find reproducible solutions as efficiently as MAP-Elites-sampling, maintaining QD-score losses close to those of MAP-Elites and significantly higher than other baselines. 
Deep-grid uses neighbours in the BD-space to approximate the true fitness and BD of a solution. However, QD-Gym tasks have a high-dimensional search space mapped to a low-dimensional BD space. Thus, the relation between genotype and BD space is probably less straightforward than in usual QD-tasks and can be confusing for this BD-neighbourhood-based mechanism.
Interestingly, MAP-Elites-ES constantly gets lower QD-score losses than all other baselines ($p<5\mathrm{e}{-7}$ except for QDHopper: $p<5\mathrm{e}{-4}$). 
As mentioned in Sec.~\ref{sec:uncertain_env}, this algorithm uses a sampling-based approximation of the gradient to direct the variation toward high-performing or diverse solutions. This approximation of the search space structure makes the variation operator less sensitive to luck. 
In addition, MAP-Elites-ES re-evaluates each solution considered for addition $30$ times, where MAP-Elites-sampling only re-evaluates them $10$ times. MAP-Elites-ES selects its offspring thoughtfully, unlike MAP-Elites-sampling, it can thus afford to spend more time on re-evaluation. MAP-Elites-sampling with $30$ re-evaluations would not even fill the grid, spending too much time on sampling the same unpromising solutions. Thanks to these two mechanisms MAP-Elites-ES finds more reproducible solutions than all other baselines, making it a strong competitor in term of reproducibility of solutions.
\name{} also finds highly reproducible solutions across all tasks, and even manage to find more reproducible solutions than MAP-Elites-sampling in QDHalfCheetah ($p<5\mathrm{e}{-5}$). Yet, \name{} does not have any explicit mechanisms to handle uncertainty in its archive-management contrary to Deep-grid, MAP-Elites-ES or MAP-Elites-sampling. It manages to find reproducible solutions while managing the archive the exact same way as MAP-Elites. 
Thus, the reproducibility of the solutions found by \name{} is probably inherent to its variation mechanisms. This result corroborates the hypothesis that PG variations bring the optimisation process to reproducible areas of the search space contrary to GA variations.

\subsection{Corrected Max-fitness and Max-fitness loss}

\name{} also outperforms all baselines according to the Corrected Max-fitness ($p<1\mathrm{e}{-9}$)
and also find significantly more reproducible solutions according to the Max-fitness loss, achieving less than $20\%$ loss across all tasks and even less than $10\%$ loss in QDWalker and $5\%$ loss in QDHalfCheetah and QDAnt. 
MAP-Elites-ES achieves similar score on QDWalker and QDAnt but higher loss on QDHalfCheetah (still less than $20\%$) and lower loss on QDHopper, while MAP-Elites-sampling achieves similar scores on QDAnt and QDHopper, but significantly higher loss in QDWalker and QDHalfCheetah. In comparison, MAP-Elites reaches significantly higher Max-fitness loss than \name{} across all tasks ($p<1\mathrm{e}{-7}$ except QDHopper: $p<2\mathrm{e}{-2}$), supporting previous results.
Deep-grid obtains really poor scores on this metric, getting even bigger loss values than MAP-Elites, especially in QDWalker. As highlighted before, Deep-grid has not been designed for such high-dimensional search spaces. Also, the original Deep-grid work has already shown that the main advantage of the Deep-grid algorithm lies in the BD-reproducibility of its solutions, and less in their fitness-performance or fitness-reproducibility. Deep-grid has a really high variance in results compared to other approaches, causing their boxplots to appear flattened on Fig.~\ref{fig:loss_stochastic}. Thus, we added the same graph without Deep-grid in Appendix~\ref{app:stochastic}, to allow to visualise the full range of variation of the other baselines.
These results illustrate again the high reproducibility of the solutions found by \name{}, that has no explicit uncertainty-handling mechanisms.

\begin{figure*}[t!]
\centering
\includegraphics[width = 0.95\hsize]{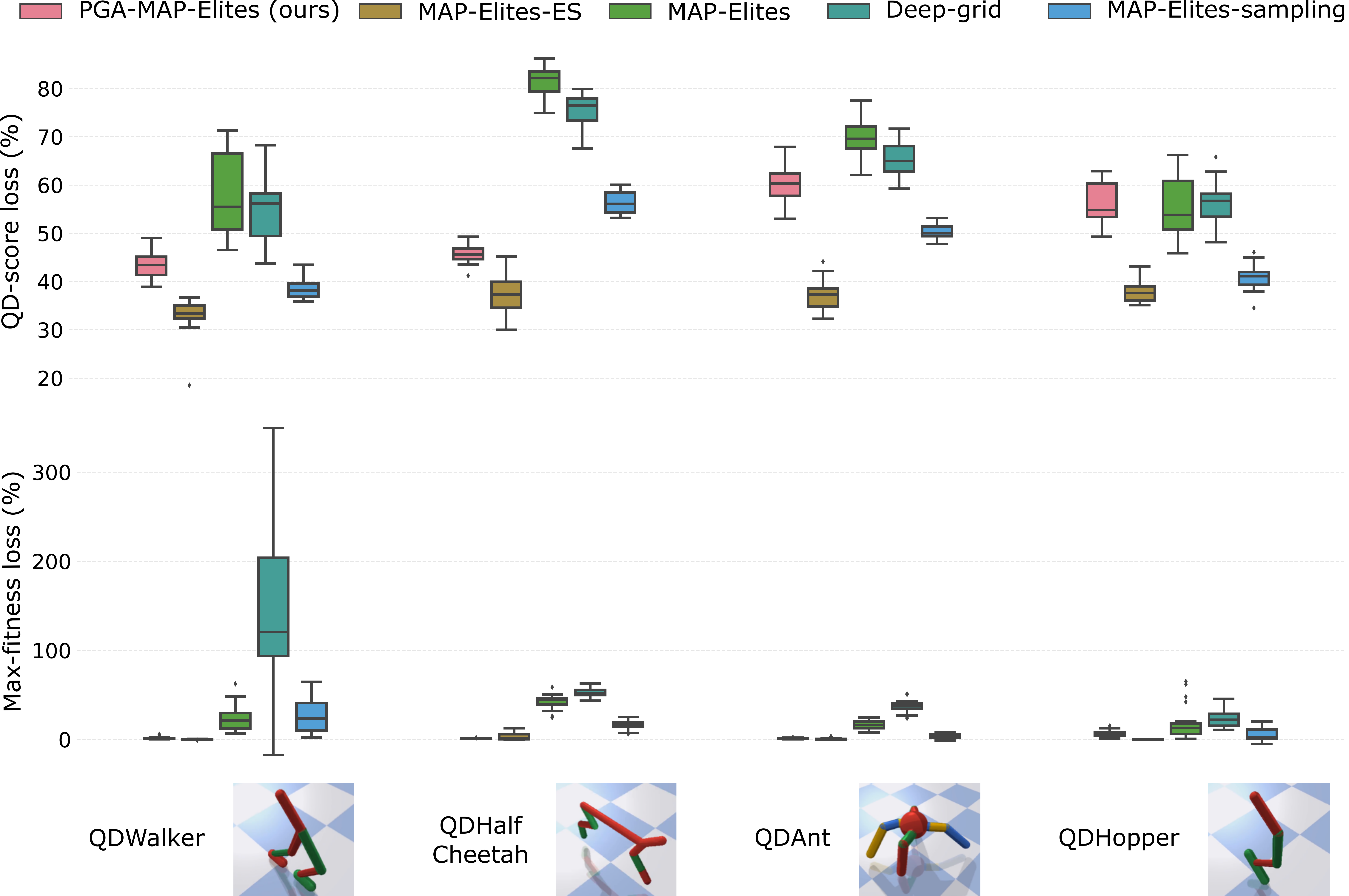}
\caption{
    \textbf{Loss comparison:}
    Comparison of the QD-score loss (top) and Max-fitness loss (bottom) due to the lack of reproducibility of the solutions. 
    To compute these losses, all solutions of the archive are replaced in a Corrected archive according to their average BD and fitness over $50$ replications. The loss corresponds to the difference of each metric in proportion of the original metric.
    Lower loss indicates better reproducibility.
    The boxes represent the distribution of the loss for a given algorithm across $20$ runs.
}
\label{fig:loss_stochastic}
\end{figure*}

\subsection{Conclusion}

\name{} is the only algorithm considered in this study, apart from vanilla MAP-Elites, that does not have any explicit mechanism to handle uncertainty. In comparison, MAP-Elites-sampling and MAP-Elites-ES both uses sampling to get a better approximate of the true BD and fitness of solutions, while Deep-grid uses neighbouring individuals for the same purpose. In this sense, \name{} is closer to vanilla MAP-Elites, and despite this similarity it proves to generate solutions that are significantly more reproducible, on top of being diverse and high-performing. 
Thus, the reproducibility of the solutions found by \name{} is inherent to its variation mechanism, based on policy-gradients.
This allows \name{} to maintain the diversity and the quality of its final archive, making it a strong QD baseline for uncertain applications. 

\section{Ablation of \name{}} \label{sec:ablation}

The core principle of \name{} is to introduce a new PG variation operator for MAP-Elites. 
At each generation, \name{} generates $b$ offsprings from parents selected from the archive.
Among these offsprings, $\lfloor{\nevo \cdot b}\rfloor$ are generated using GA, and $\lceil{(1-\nevo) \cdot b}\rceil$ using PG. The last of the PG-generated solutions is a copy of the greedy actor of the critics. 
In this section, we propose an ablation study to better understand the impact of these variation operators.

\subsection{Impact of the proportion of PG and GA variation}

We first study the impact of the proportion of PG and GA variations: $\nevo$.

\subsubsection{Baselines:}
In the above experiments, the implementation of \name{} uses $\nevo = 0.5$ and $b=100$. In other words, $50$ offsprings are generated through GA and $50$ through PG, among which $1$ is a copy of the greedy actor. 
In this section, we compare multiple versions of \name{} with $b=100$ but different values of the $\nevo$ parameter:
\begin{itemize}
    \item $\nevo = 1$ - corresponds to the standard MAP-Elites algorithm with only GA variations.
    \item $\nevo = 0.75$ - corresponds to $75\%$ of GA and $25\%$ of PG. 
    \item $\nevo = 0.5$ - is the standard \name{} algorithm.
    \item $\nevo = 0.25$ - corresponds to $25\%$ of GA and $75\%$ of PG. 
    \item $\nevo = 0$ - corresponds to full PG variations. Importantly, it does not correspond to TD3. First, \name{} with $\nevo = 0$ selects the parents from the archive, while TD3 does not perform any selection and always modifies the greedy actor. Second, in \name{} with $\nevo = 0$, the training of the greedy and the variations are applied only every generation. In other word, actors are modified only between episodes, while in TD3 the greedy actor is trained within episode. 
\end{itemize}

\begin{figure*}[!t]
\centering
\includegraphics[width = 0.98\hsize]{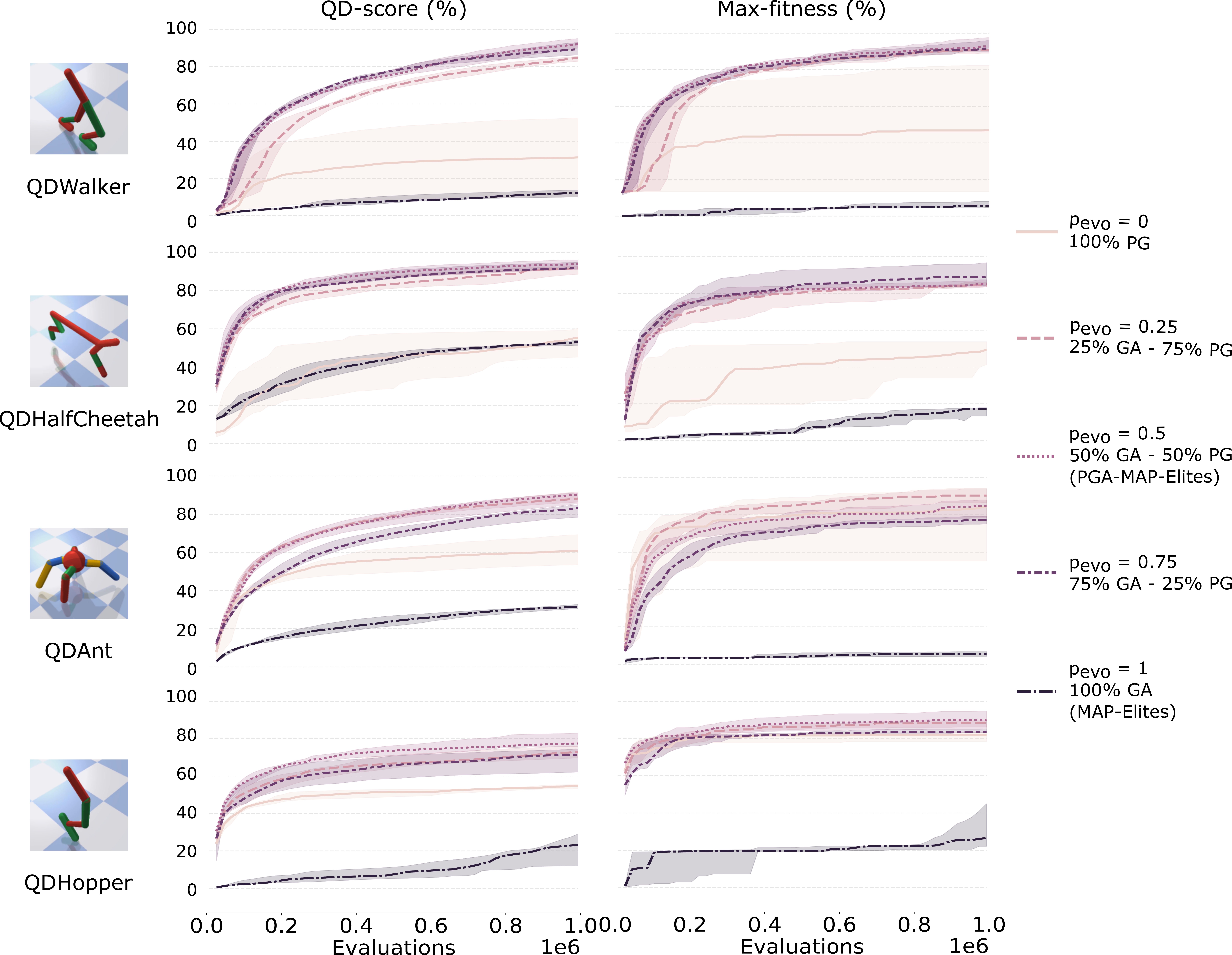}
\caption{
    \textbf{Ablation $\nevo$ - Main results:}
    Comparison of the QD-score (left) and Max-fitness (right) of \name{} with different mutation proportions on the QDGym tasks. Each experiment is replicated $10$ times, the solid line gives the median over replications and the shaded area to the first and third quartiles.
}
\label{fig:progress_ablation}
\end{figure*}

\subsubsection{Main results:}
Fig.~\ref{fig:progress_ablation} summarises the results of this study, the corresponding archives are given in Appendix~\ref{app:ablation}. All runs are replicated $20$ times.
\name{} with $\nevo = 1$, which corresponds to MAP-Elites, under-performs \name{} with $\nevo = [0.25, 0.5, 0.75]$: its QD-score is order-of-magnitude lower ($p<5\mathrm{e}{-4}$) as well as its Max-fitness ($p<5\mathrm{e}{-4}$). 
This first result illustrates the importance of policy-gradients in \name{}.
On the opposite side of the spectrum, \name{} with $\nevo = 0$, corresponding to PG-only, also reaches significantly lower QD-score ($p<5\mathrm{e}{-4}$) than $\nevo = [0.25, 0.5, 0.75]$. 
This variant manages to find the same max-fitness solutions as the other baselines in QDAnt and QDHopper, but not in QDWalker ($p<5\mathrm{e}{-3}$) and QDHalfCheetah ($p<5\mathrm{e}{-4}$).
This second result highlights the importance of the divergent fitness-agnostic search of the GA variation operator.

The last three variants: $\nevo = [0.25, 0.5, 0.75]$ perform equally across tasks and metrics. 
The ideal value of $\nevo$ seems to vary with the task. 
$\nevo = 0.25$, which corresponds to \name{} with predominant PG, converges a bit slower than the two others and even reaches a lower value of QD-score in the QDWalker task ($p<5\mathrm{e}{-2}$). However, it performs similarly to the others in the QDAnt task.
On the contrary, $\nevo = 0.75$, \name{} with predominant GA, performs similarly to the others in QDWalker and QDHalfCheetah but reaches a lower QD-score in QDAnt ($p<1\mathrm{e}{-2}$). 
These results illustrate the importance of balancing GA and PG variations in \name{} and indicate that the algorithm is quite robust to the choice of this hyperparameter as long as it does not take extreme values. Overall, $\nevo = 0.5$, which corresponds to the value chosen for \name{} in the previous sections, shows the best trade-off.

\begin{figure*}[t!]
\centering
\includegraphics[width = 0.98\hsize]{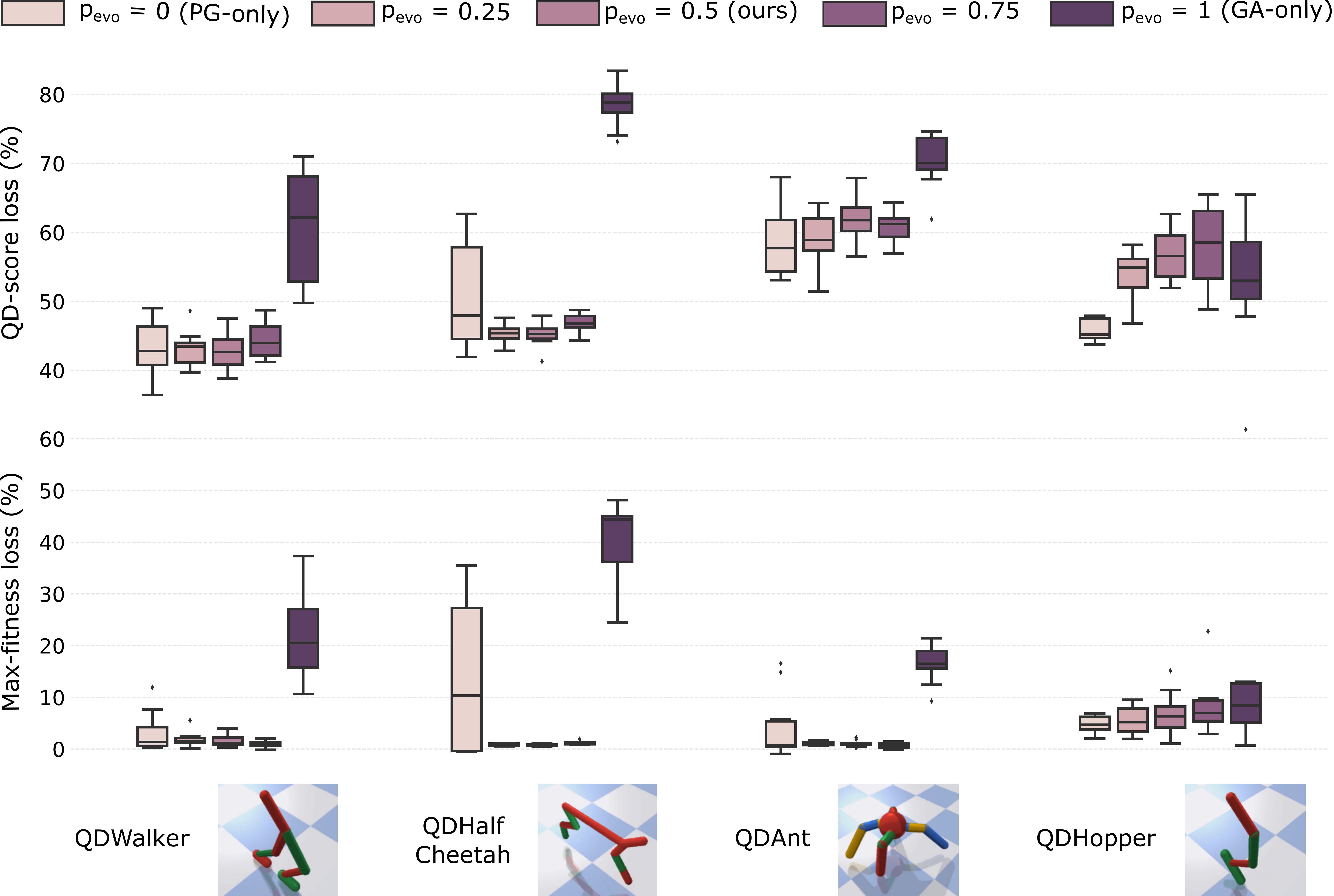}
\caption{
    \textbf{Ablation $\nevo$ - Loss comparison:}
    Comparison of the QD-score loss (top) and Max-fitness loss (bottom) due to the lack of reproducibility of the solutions, for \name{} with different variation operator proportions. 
    To compute these losses, all solutions of the archive are replaced in a Corrected archive according to their average BD and fitness over $50$ replications. The loss corresponds to the difference of each metric in proportion of the original metric.
    Lower loss indicates better reproducibility.
    The boxes represent the distribution of the loss for a given algorithm across $10$ runs.
}
\label{fig:loss_ablation_stochastic}
\end{figure*}

\subsubsection{Reproducibility results:} Fig.~\ref{fig:loss_ablation_stochastic} gives the reproducibility analysis for this ablation. The corresponding Corrected metrics and Corrected archives are given in Appendix~\ref{app:ablation}.
These results draw similar conclusions to previous results of this section but also to those obtained in Sec.~\ref{sec:stochastic}. 
Overall, all algorithms are impacted by the re-evaluation mechanisms but \name{} with $\nevo{}=[0.25, 0.5, 0.75]$ still outperform the extreme values both in term of Corrected metrics and Loss metrics.
Again, $\nevo{}=[0.25, 0.5, 0.75]$ are equivalent and $\nevo{}=0.5$, the value chosen earlier in the paper, shows the best trade-off.
Fig.~\ref{fig:loss_ablation_stochastic} shows that \name{} with $\nevo{}=1$, which corresponds to MAP-Elites, is more impacted than the others by the re-evaluation process. As all other algorithms rely on critics approximations, this observation corroborates the hypothesis that the robustness mechanism enforced by the PG variation via the TD3 training allows to better handle uncertainty. 
The Coverage and QD-score loss of \name{} with $\nevo{}=0$, which corresponds to PG variations only, are equivalent or even lower in average to those of other baselines, supporting the hypothesis of the importance of critic-based variation for the reproducibility of solutions.
However, it displays a greater variance across runs, which is coherent with the high variance observed in the results in Fig.~\ref{fig:progress_ablation}. \name{} with $\nevo{}=0$ might lack exploration compared to the other algorithms which all use GA variation, which enhance the difference in performance between lucky and unfortunate runs.

\begin{figure*}[t]
\centering 
\includegraphics[width = 0.98\hsize]{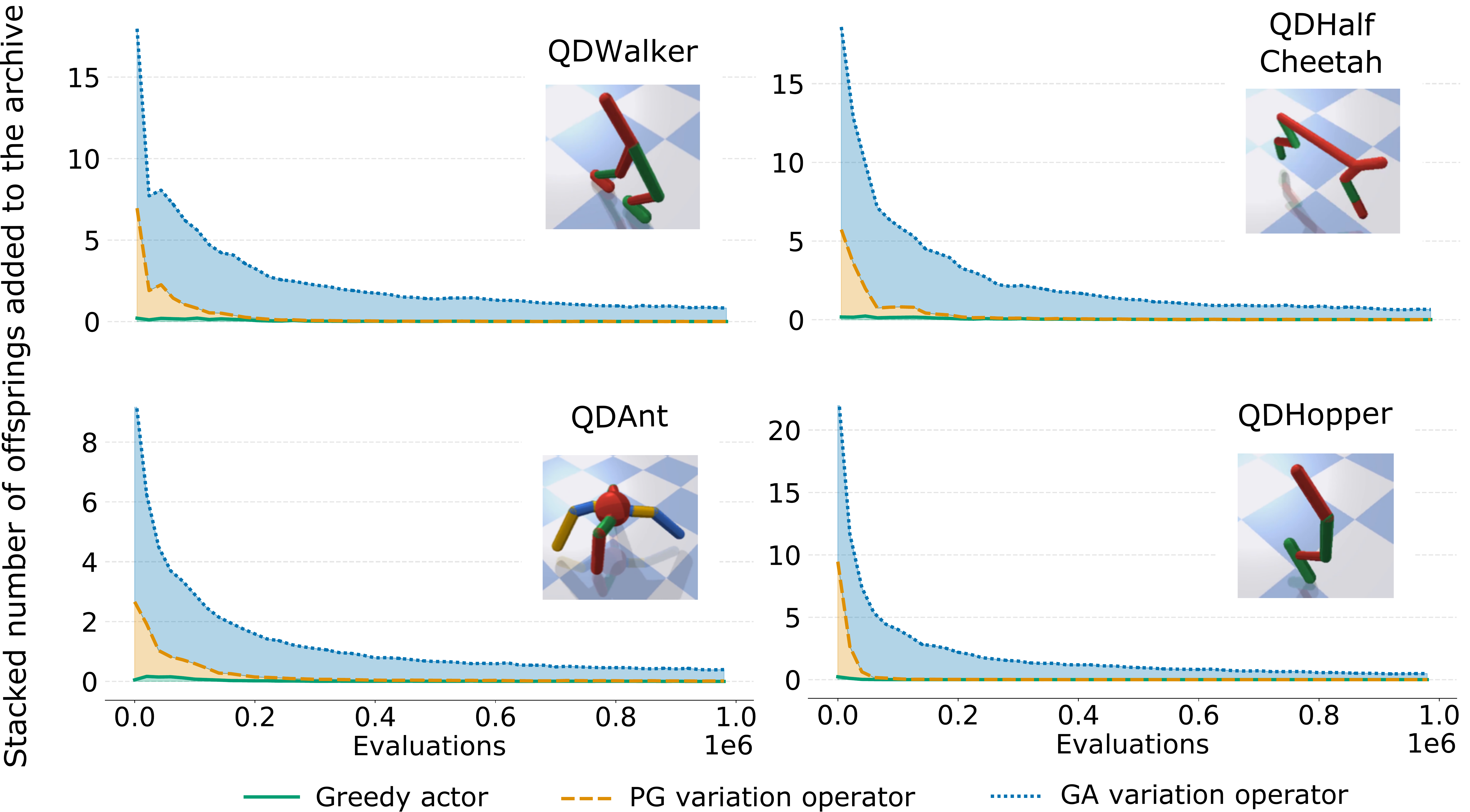}
\caption{
    \textbf{Variation operators contribution:}
    Stacked contribution of each variation operator of \name{}: GA, PG and Greedy. The number of offspring generated by each operator is fixed: $50$ for GA, $49$ for PG and $1$ for Greedy. This graph gives the number of these offspring added to the archive for each operator. Offspring are added to the archive when they either discover a new cell or improve the existing elite of their cell. The results are averaged over 20 replications of \name{}, and smoothed on $2000$ generations ($2*10^4$ evaluations) to increase the readability. They are stacked over operators to facilitate comparison. 
}
\label{fig:variation_usage}
\end{figure*}

\subsection{Variation operators contribution}

This section studies the contribution to archive improvement of each variation operator.

\subsubsection{Metrics}

We consider three operators: \textbf{GA variation}, \textbf{PG variation} and \textbf{Greedy actor}.
The first two correspond to the two variation operators of \name{}. \textbf{Greedy actor} refers to a copy of the actor associated with the critics. It is considered among the offspring of each generation.
We propose to visualise how each of these operators improves the archive across time by counting the number of solutions added to the archive it generates. We represent the median of this quantity across 20 runs against the number of evaluations.

\subsubsection{Results} 

Fig.~\ref{fig:variation_usage} summarises this analysis.
While we observed in previous sections that the PG variation operator is necessary to get good results, here we can see that it is only relevant during the early stage of the process. On the contrary, the GA operator contributes to archive improvement throughout the process. This result suggests that the PG operator is used to find the region of the search space containing high-performing solutions: the hyper-volume of Elites, as introduced by \citet{vassiliades2018discovering}. The GA operator is then used to exploit this hyper-volume.

Across all tasks, the total number of offspring added to the archive per generation reaches its maximum at the start of the process. These values support previous analysis showing that interesting solutions are harder to find for QDAnt and easier to find in QDHopper. 
All tasks show a first phase during which the three operators improve the archive.
During this phase, the \textbf{PG variation} and the \textbf{GA variation} have an equivalent impact. The former guide the optimisation process toward promising regions of the search space, and the latter explore these regions. 
Throughout this first phase, the \textbf{Greedy actor}, also guided by the critics toward promising regions, is often added to the archive.
This phase is of equivalent length for QDWalker and QDHalfCheetah, but longer for QDAnt, and significantly shorter for QDHopper.
During the rest of the process, the impact of all variations decreases. The \textbf{PG variation} is the most impacted and adds almost no new solution to the archive, whereas the \textbf{GA variation} continues to discover new diverse solutions in the hyper-volume of elites found thanks to the critics. During this second phase, the \textbf{Greedy actor} keep on being added to the archive episodically but less frequently.
Thus, despite being essential to finding promising solutions in the QD-Gym domains, the PG variation operator is only crucial for the first phase of the process during which it guides the optimisation process toward the hyper-volume of elites, that is then extensively explored by the GA operator. 
These results corroborate the previous ones, showing the importance of both PG and GA variation operators in \name{}.

    
    
    
    
    
    

\section{Conclusions and Discussion}

This work proposes an in-depth analysis of \name{}, a new variant of MAP-Elites that allows to successfully apply the QD approach to uncertain environments with high-dimensional search space. 
\name{} extends MAP-Elites with a new DRL-inspired variation operator that leverages the estimation of critic networks and pairs it with the standard GA variation operator for exploration. 
We first prove that \name{} significantly outperforms existing methods on four robotic locomotion tasks, demonstrating its ability in uncertain tasks with high-dimensional search-space. We also show that these results hold in deterministic variants of the same tasks.
We thus secondly propose an analysis of the reproducibility of the solution generated by \name{} in such uncertain domains. We demonstrate that these solutions are highly reproducible, approaching the solutions generated by QD baselines designed specifically for uncertain applications.
Thirdly, we study in detail the impact of the proportion of PG and GA variations on \name{}. We demonstrate the importance of the two operators to preserve both the performance and the reproducibility of the solutions. We also show that the performance of \name{} are robust to the proportion of PG and GA, as long as the two operators are present. 
Finally, we analyse the contribution of each variation operator to archive improvement. The results indicate that the PG operator is crucial at the beginning of the optimisation process to discover the hyper-volume of elites containing the performing individuals. The GA variation operator remains essential throughout the process to guarantee sufficient exploration of the search space. 
Overall, these results highlight the importance of policy-gradients in \name{} and prove their potential to improve MAP-Elites and allow to apply QD to evolve diverse high-quality DNN controllers.

These analyses point toward possible directions for future work. The ablation study indicates that the PG variation is key during the first phase of the optimisation process, where it contributes to improving the collection of solutions. During this phase, it seems to guide the process toward promising regions of the search space that are then explored by the GA. Thus, an interesting direction of research would be to propose an adaptive mechanism for the proportion of GA and PG as variation operators. 
Also, this work proposes an analysis of \name{} applied to robotic locomotion tasks. It would be interesting for future work to study its performance in other domains, possibly uncertain and controlled using high-dimensional DNN controllers, for example manipulation tasks.


\bibliographystyle{ACM-Reference-Format}
\bibliography{sample-bibliography}

\appendix

\newpage

\section{Algorithms Hyperparameters} \label{app:hyperparam}

\name{}, MAP-Elites, TD3, MAP-Elites-sampling and Deep-grid are all implemented in a common repository detailed in Sec.~\ref{sec:hyperparams}. The corresponding hyperparameters are summarised below ($\#\mathcal{A}$ refers to the number of actions in a task).
MAP-Elites-ES and QD-PG are not included in this table as we used the implementation and hyperparameters of the original authors.

\textsc{\begin{table}[ht]
  \label{tab:hyperparam}
  \begin{tabular}{l c c c}
    \toprule
    Parameter & MAP-Elites variants & PGA & TD3\\
    \midrule
    Nr. of random init. ($I$) & $500$ episodes & $500$ episodes & $2500$ timesteps\\
    Evaluation batch size ($b$) & $100$ & $100$ & $1$ \\
    Actor networks & [128, 128, $\#\mathcal{A}$] & [128, 128, $\#\mathcal{A}$] & [128, 128, $\#\mathcal{A}$]\\
    Critic networks & / & [256, 256, 1] & [256, 256, 1]\\
    Training batch size ($N$) & / & $256$ & $256$  \\
    Critic training steps ($n_{crit}$) & / & $300$ & / \\
    Actor training steps ($n_{act}$) & / & $50$ & / \\
    Critic learning rates ($lr_{crit}$) & / & $3\times 10^{-4}$ & $3\times 10^{-4}$ \\
    Actor learning rate ($lr_{act}$) & / & $0.005$ & $3\times 10^{-4}$ \\ 
    Replay buffer max. size & / & $10^6$ & $10^6$  \\
    Discount factor ($\gamma$) & / & $0.99$ & $0.99$  \\
    Exploration noise ($\sigma_a$) & / & $0.2$ & $0.2$  \\
    Target update freq. ($d$) & / & $2$ & $2$ \\
    Target update rate ($\tau$) & / & $0.005$ & $0.005$ \\
    Smoothing noise var. ($\sigma_p$) & / & $0.2$ & $0.2$  \\
    Smoothing noise clip ($c$) & / & $0.5$ & $0.5$  \\
    GA variation param. 1 ($\sigma_1$) & $0.005$ & $0.005$ & / \\
    GA variation param. 2 ($\sigma_2$) & $0.05$ & $0.05$ & / \\ 
  \bottomrule
\end{tabular}
\end{table}}

\newpage 

\section{Reproducibility in uncertain environment - complementary} \label{app:stochastic}

\begin{figure*}[h]
\centering
\includegraphics[width = 0.98\hsize]{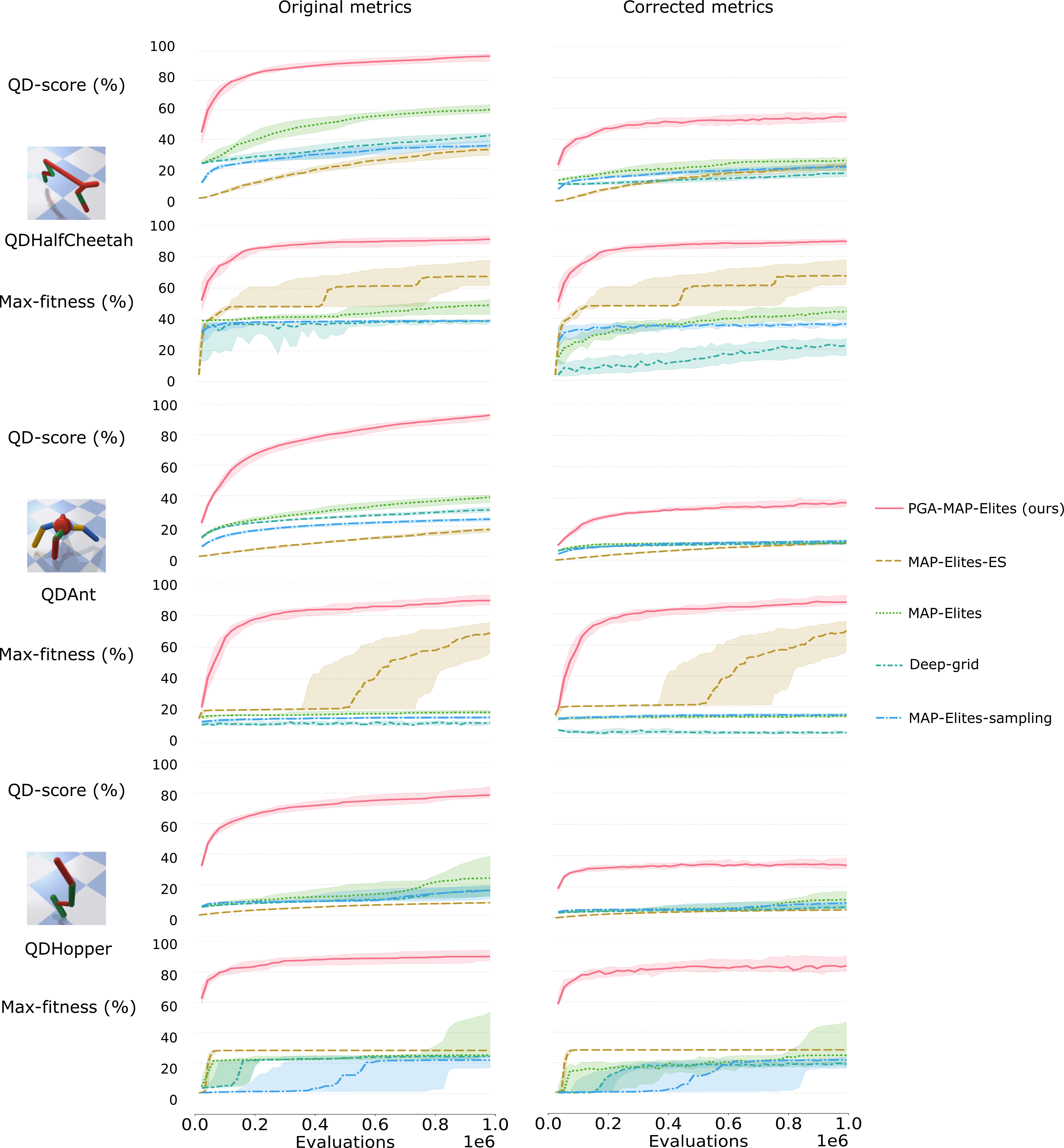}
\caption{
\textbf{Corrected results:}
Comparison of the original (left) and corrected (right) QD-score (top) and Max-fitness (bottom) of all algorithms on QDWalker.
To compute the Corrected metrics, all solutions of the archive are replaced in a Corrected archive according to their average BD and fitness over $50$ replications.
Each experiment is replicated $20$ times, the solid line corresponds to the median over replications and the shaded area to the first and third quartiles. 
}
\label{fig:progress_stochastic}
\end{figure*}

\begin{figure*}[h]
\centering
\includegraphics[width = 0.98\hsize]{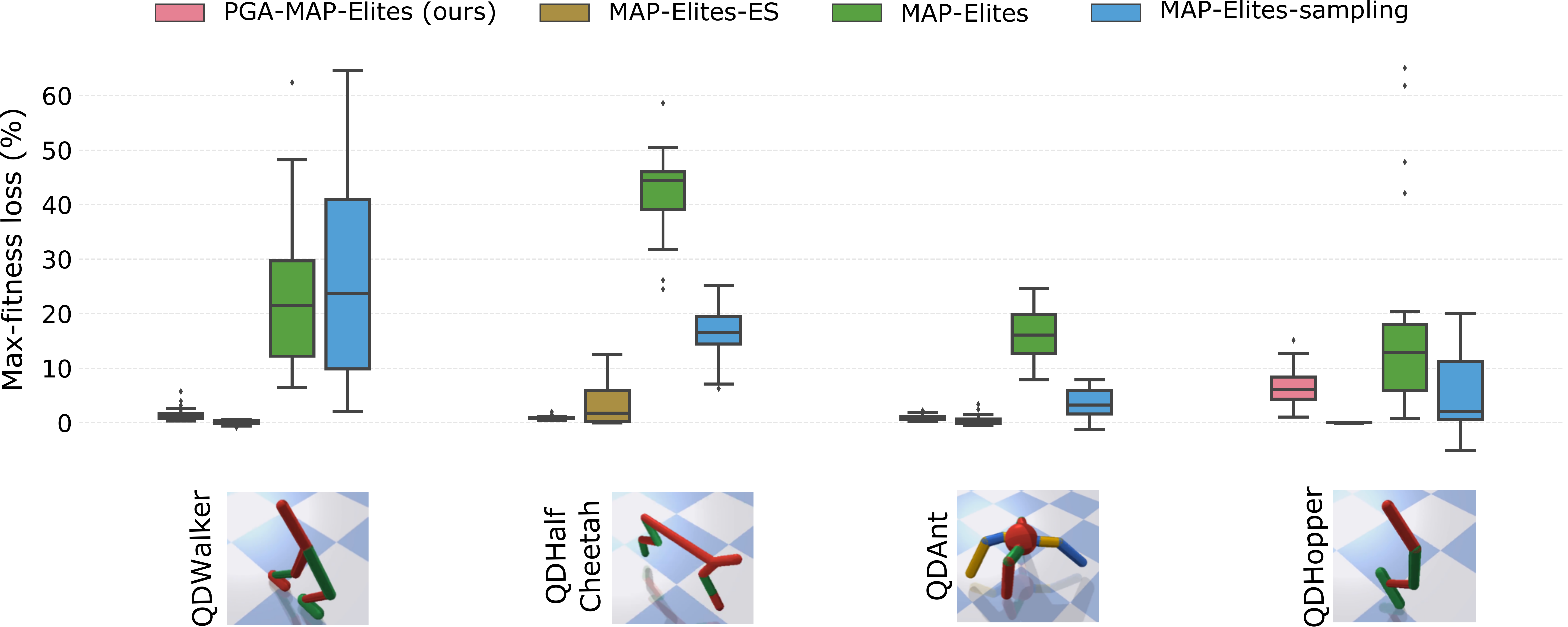}
\caption{
\textbf{Loss comparison:}
    Comparison of the Max-fitness loss due to the lack of reproducibility of the solutions. 
    To compute these losses, all solutions of the archive are replaced in a Corrected archive according to their average BD and fitness over $50$ replications. The loss corresponds to the difference of each metric in proportion of the original metric.
    Lower loss indicates better reproducibility.
    The boxes represent the distribution of the loss for a given algorithm across $20$ runs.
}
\label{fig:loss_stochastic_nodeep}
\end{figure*}

\begin{figure*}[h]
\centering 
\includegraphics[width = 0.98\hsize]{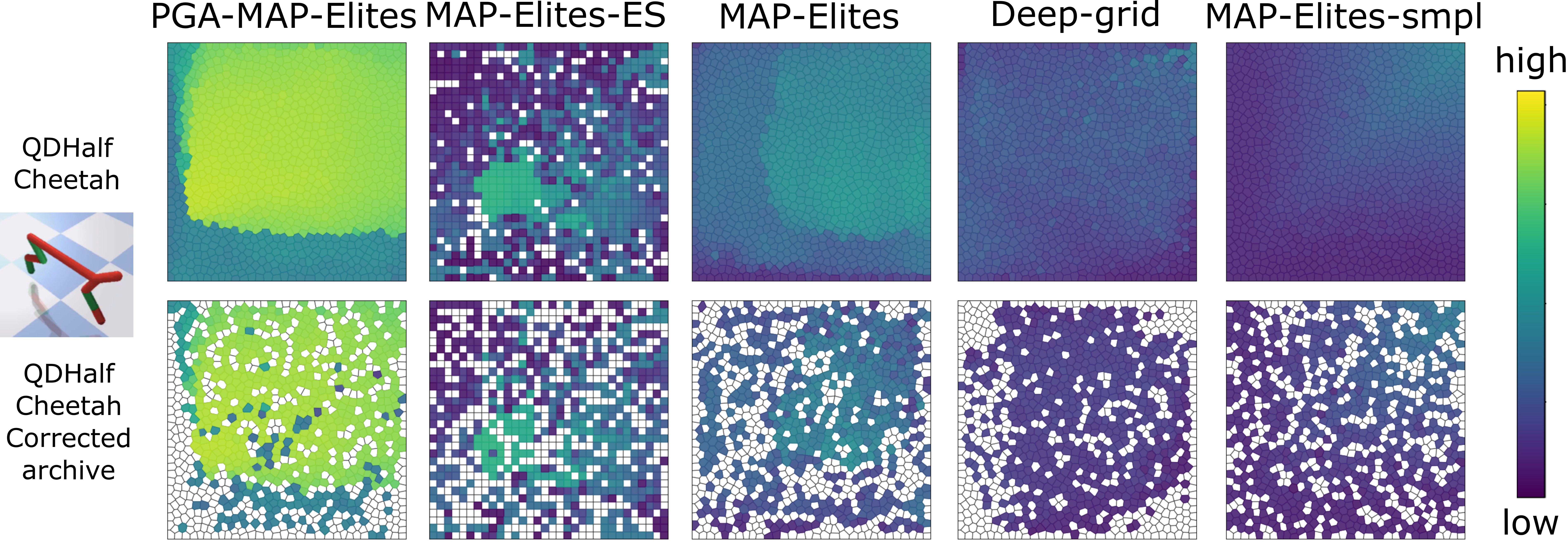}
\caption{
\textbf{Ablation $\nevo$ - Corrected archives:}
Final archives (top) and final Corrected archives (bottom) found by all algorithms in the QDHalfCheetah tasks after $10^6$ controller evaluations. The BDs in these tasks are 2-dimensional, giving a square grid. Increasing feet contact time from left to right and from bottom to top. The colour of each cell represents the fitness of the controller it contains (the lighter the better), grey colour corresponds to an empty cell. 
The Corrected archive is filled with the controllers from the standard archive with the same archive-addition rules but using the average fitness and BD over $50$ replications.
}
\label{fig:maps_stochastic}
\end{figure*}

This section complements the results displayed in Fig.~\ref{fig:progress_stochastic_part} and Fig.~\ref{fig:maps_stochastic_part}, and analysed in Sec.~\ref{sec:stochastic}.
Fig.~\ref{fig:progress_stochastic} shows the original and corrected metrics for all algorithms across remaining tasks. Fig.~\ref{fig:maps_stochastic} shows the corresponding archives for the QDHalfCheetah tasks. 
We also display the Max-fitness loss for all baselines except Deep-grid in Fig.~\ref{fig:loss_stochastic_nodeep}

\section{Ablation-study - complementary} \label{app:ablation}

\subsection{Archives}

\begin{figure*}[h]
\centering 
\includegraphics[width = 0.98\hsize]{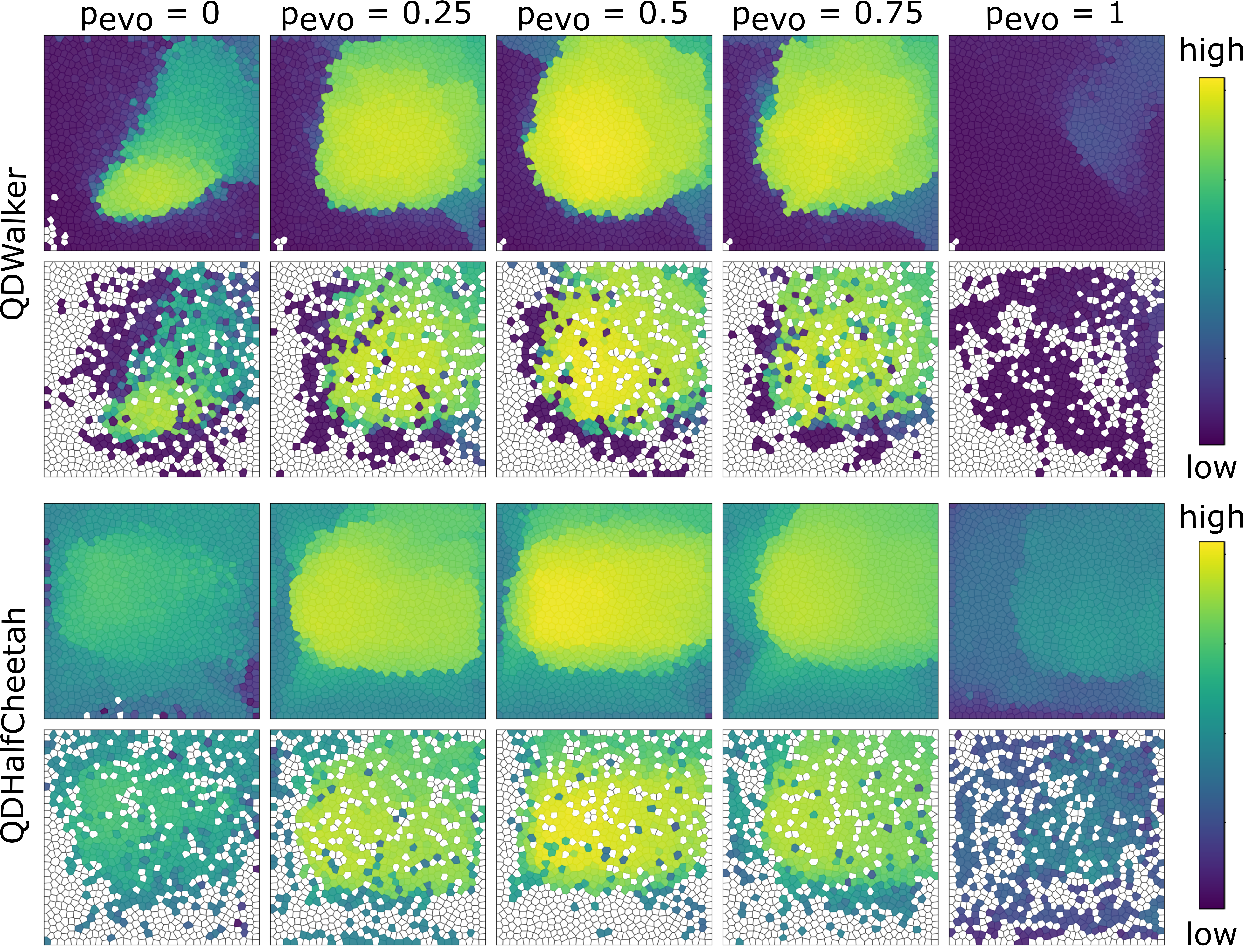}
\caption{
\textbf{Ablation $\nevo$ - Corrected archives:}
Final archives (top line) and final corrected archives (bottom line) found by \name{} with different mutation proportion in the QDWalker (top) and QDHalfCheetah (bottom) tasks after $10^6$ controller evaluations. The BD in these tasks are 2-dimensional, giving a square grid. Increasing feet contact time from left to right and from bottom to top. The colour of each cell represents the fitness of the controller it contains (the lighter the better), grey colour corresponds to an empty cell. 
The Corrected fitness and BD of each controller is computed as the average over $50$ replications of this controller. The Corrected archive is filled with the same archive rules but using these corrected scores. 
}
\label{fig:maps_ablation}
\end{figure*}

Fig.~\ref{fig:maps_ablation} display the archives and corrected archives obtained for one run of each baseline considered in the ablation study in Sec.~\ref{sec:ablation}.
These archives support the results from Sec.~\ref{sec:ablation} and show that $\nevo{}=0.25$, $\nevo{}=0.5$ and $\nevo{}=0.75$ outperform the two extreme baselines $\nevo{}=0$, only PG, and $\nevo{}=1$, only GA. This highlights again the complementary of PG and GA variation in \name{}. 
However, the archives given in Fig.~\ref{fig:maps_ablation} only represent one run and should be compared gingerly, especially for \name{} with $\nevo{}=0$ which has a really high variance across runs. 

\subsection{Reproducibility in uncertain environment}

\begin{figure*}
\centering
\includegraphics[width = 0.98\hsize]{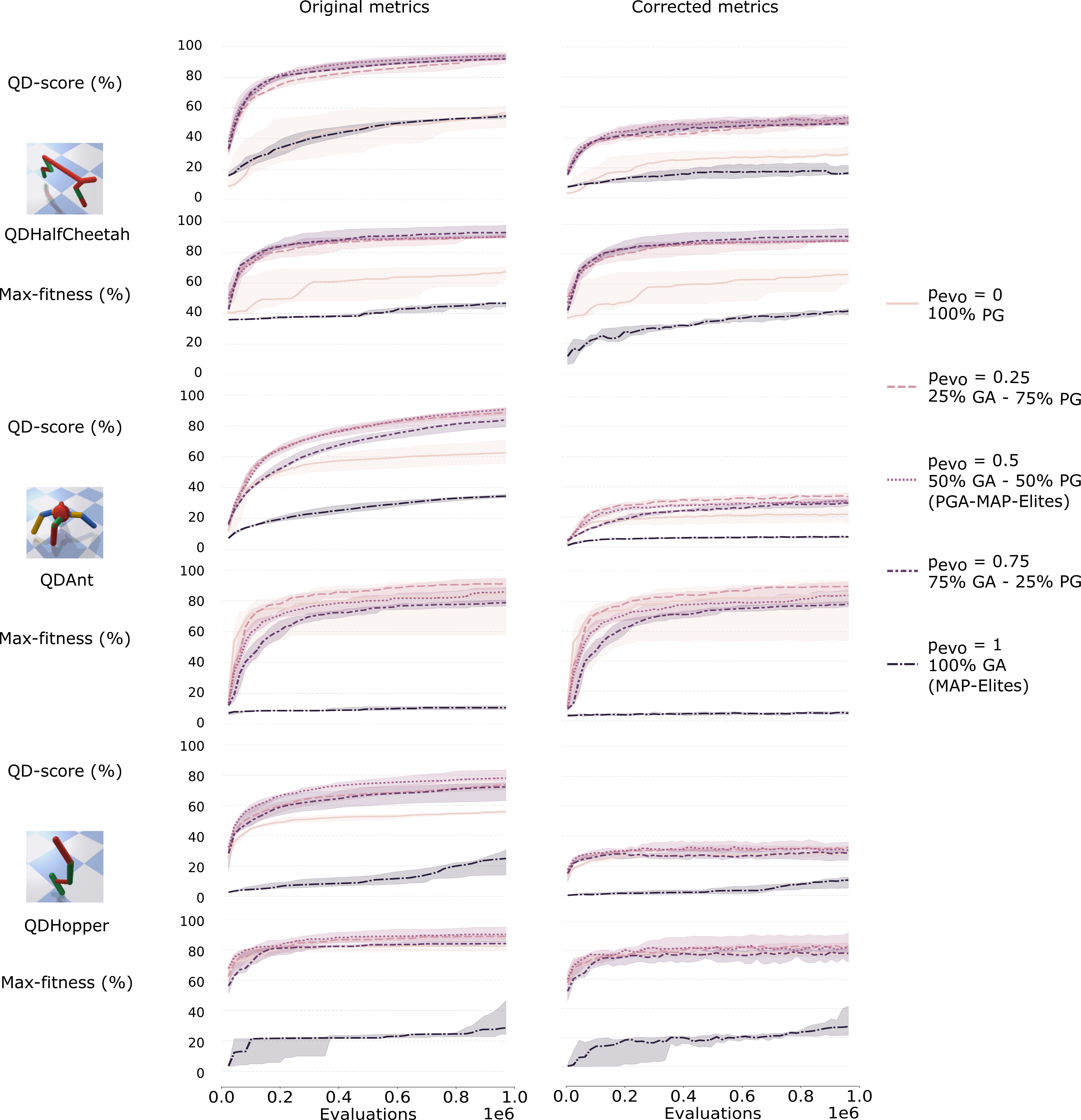}
\caption{
\textbf{Ablation $\nevo$ - Corrected results:}
Comparison of the original (left) and Corrected (right) QD-score (top) and Max-fitness (bottom) of \name{} with different mutation proportion on each QDGym task.
To compute the Corrected metrics, all solutions of the archive are replaced in a Corrected archive according to their average BD and fitness over $50$ replications.
Each experiment is replicated $10$ times, the solid line corresponds to the median over replications and the shaded area to the first and third quartiles. 
}
\label{fig:progress_ablation_stochastic}
\end{figure*}

We also give the Corrected metrics complementing the ablation study of Sec.~\ref{sec:ablation} in Fig.~\ref{fig:progress_ablation_stochastic}. 

\section{Coverage comparison} \label{app:coverage}

\subsection{Performance of \name{}}

\begin{figure*}[h]
\centering 
\includegraphics[width = 0.98\hsize]{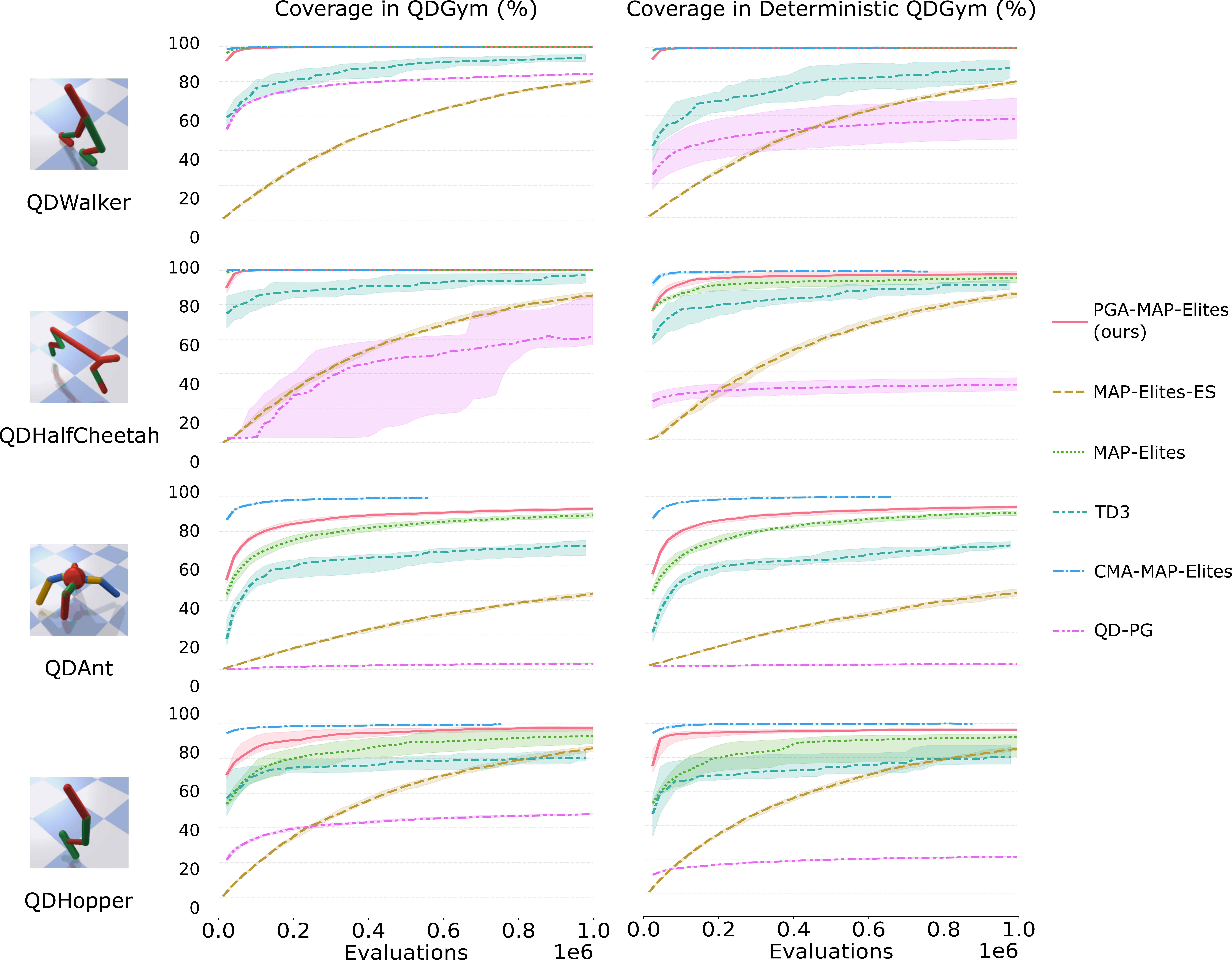}
\caption{
    \textbf{QDGym and Deterministic QDGym results:}
    Coverage of all algorithms on each QDGym (left) and Deterministic QDGym (right) task for $10^6$ controller evaluations. Each experiment is replicated $20$ times, the solid line corresponds to the median over replications and the shaded area to the first and third quartiles.
}
\label{fig:progress_coverage}
\end{figure*}

We display in Fig. \ref{fig:progress_coverage} the Coverage metrics corresponding to Section \ref{sec:deterministic} in both uncertain and deterministic environment.
This additional metric shows that \name{} also retains the divergent search capability of QD-algorithms. 
It achieves coverages similar to MAP-Elites and CMA-MAP-Elites for the QDWalker and QDHalfCheetah tasks and even outperforms MAP-Elites in QDHopper ($p<1\mathrm{e}{-2}$) and QDAnt ($p<5\mathrm{e}{-8}$). 
In this last task, CMA-MAP-Elites manages to reach a better coverage than all the other baselines within only $2*10^5$ evaluations ($p<1\mathrm{e}{-6}$) but longer run-time. As mentioned before, QDAnt likely has a more complex subspace of promising policies than other QD-Gym tasks, making it harder to discover and explore, which might explain the good performance of CMA-MAP-Elites in this domain.
\name{} also outperforms TD3 on the Coverage metric ($p<5\mathrm{e}{-5}$). Interestingly, TD3, which is not a QD algorithm, manages to get relatively good coverage of the BD space, in particular in QDWalker and QDHalfCheetah. However, it does not optimise these local solutions and only focuses on the best-performing one, as highlighted by its poor QD-score. 
\name{} also performs better than QD-PG and MAP-Elites-ES, for causes similar to those of the QD-Score. 
These results are similar in both versions of the environments. Only QD-PG obtains significantly lower coverage when applied to Deterministic QD-Gym tasks, probably due to the random initialisation of the robot encouraging exploration.

\subsection{Reproducibility in uncertain environments}

\begin{figure*}[h]
\centering
\includegraphics[width = 0.98\hsize]{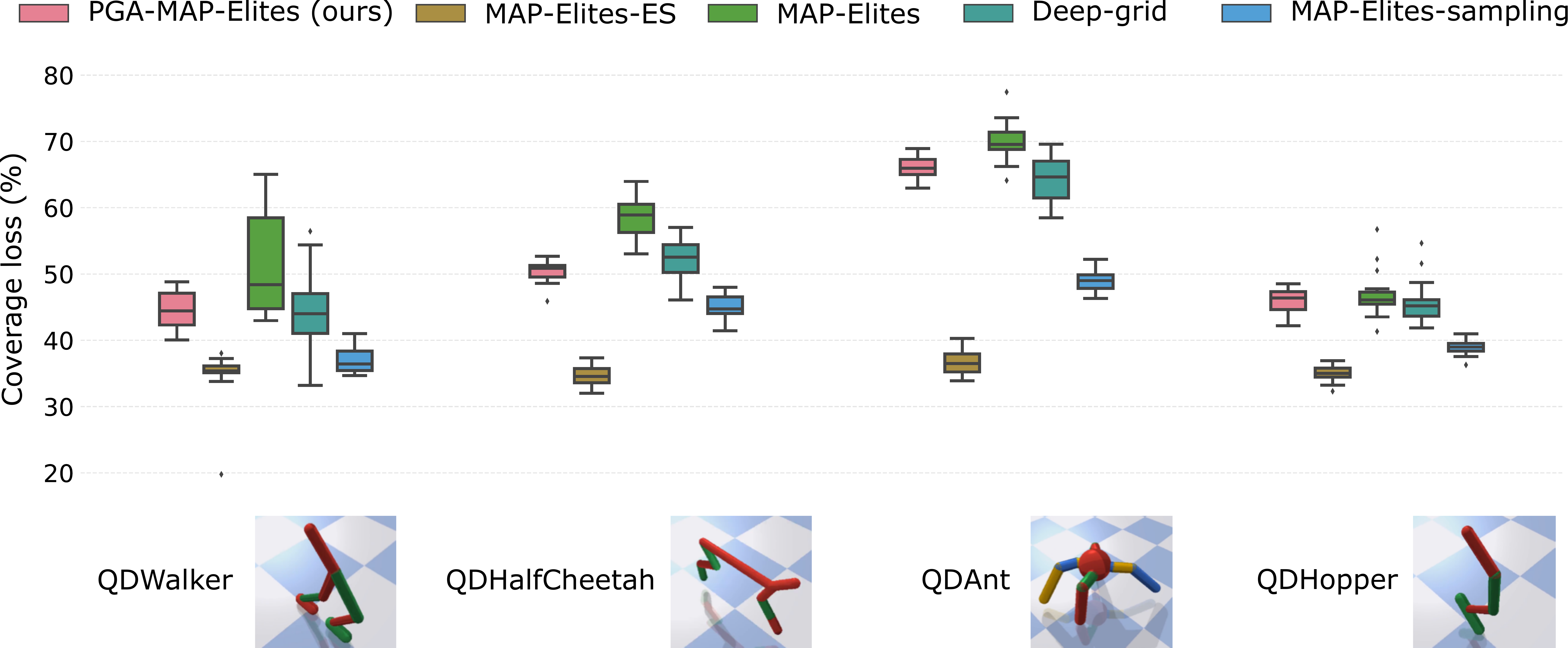}
\caption{
    \textbf{Loss comparison:}
    Comparison of the Coverage loss.
    To compute these losses, all solutions of the archive are replaced in a Corrected archive according to their average BD and fitness over $50$ replications. The loss corresponds to the difference of each metric in proportion of the original metric.
    Lower loss indicates better reproducibility.
    The boxes represent the distribution of the loss for a given algorithm across $20$ runs.
}
\label{fig:loss_coverage}
\end{figure*}

Fig. \ref{fig:loss_coverage} gives the Coverage loss corresponding to Section \ref{sec:stochastic}.
According to the metrics in Appendix~\ref{app:stochastic}, MAP-Elites-sampling and MAP-Elites-ES have a higher diversity in their final Corrected archive than all other algorithm in QDWalker ($p<1\mathrm{e}{-5}$) and QDHopper ($p<1\mathrm{e}{-7}$), and MAP-Elites-sampling still does in QDHalfCheetah ($p<5\mathrm{e}{-4}$). However, in QDAnt they are performing equivalently to \name{} and Deep-grid.
These results, as well as the archives displayed in Fig.~\ref{fig:maps_stochastic}, indicate that MAP-Elites-sampling and MAP-Elites-ES are the only algorithms that manage to discover reproducible solutions all over the BD-space, but their sample inefficiency makes them less efficient in QDAnt.
The Coverage loss results in Fig.~\ref{fig:loss_coverage} are similar to those of QD-score loss, and the comparison across algorithms stay alike: 
MAP-Elites-ES is the algorithm that finds the more reproducible solutions across all tasks ($p<1\mathrm{e}{-7}$ except in QDHalfCheetah: $p<5\mathrm{e}{-5}$), and MAP-Elites-sampling stays second in this category ($p<5\mathrm{e}{-7}$ except in QDHalfCheetah: $p<5\mathrm{e}{-4}$). Their sampling mechanism increases their chance to store solutions in more accurate cells than MAP-Elites or \name{} which only rely on one evaluation to estimate the BD of solutions. In the case of MAP-Elites-ES, the sampling-based estimate of the gradient probably also guide variation toward reproducible solutions.
Despite also being designed for uncertain applications, Deep-grid obtains higher Coverage-loss than these two baselines ($p<5\mathrm{e}{-7}$ except in QDHalfCheetah: $p<5\mathrm{e}{-4}$). As highlighted before, its neighbours-based mechanism is not built for such high-dimensional search spaces. 
Among all baselines, MAP-Elites observes the higher performance drop ($p<5\mathrm{e}{-5}$ except in QDHalfCheetah: $p<6\mathrm{e}{-3}$), except on QDHopper where it performs similarly to \name{} and Deep-grid.

The coverage-losses of \name{} are close to those of the three uncertainty-handling QD algorithms, staying significantly lower than those of MAP-Elites ($p<5\mathrm{e}{-5}$ and in QDHalfCheetah $p<3\mathrm{e}{-3}$) except on QDHopper.
These results lead to the same conclusions as for the QD-score loss metric.
However, interestingly, the solutions found by \name{} are slightly less reproducible in terms of Coverage than they were in terms of QD-score: \name{} obtains slightly higher Coverage loss than QD-score loss in QDHalfCheetah and QDAnt. 
On the contrary, MAP-Elites solutions are more reproducible in terms of Coverage than QD-score (even if they stay less reproducible than those of all other baselines). 
As mentioned earlier, MAP-Elites and \name{} use the same archive-management and differ only in variation operators. 
Thus, this difference could be related to critic-based variations: as the critic is only aware of fitness values and not BD, it may encourage reproducibility in terms of fitness but not directly in terms of BD. However, as reproducibility in terms of fitness and BD are likely correlated, we still observe a better Coverage-loss for \name{}.

\subsection{Ablation of \name{}}

\begin{figure*}[h!]
\centering
\includegraphics[width = 0.65\hsize]{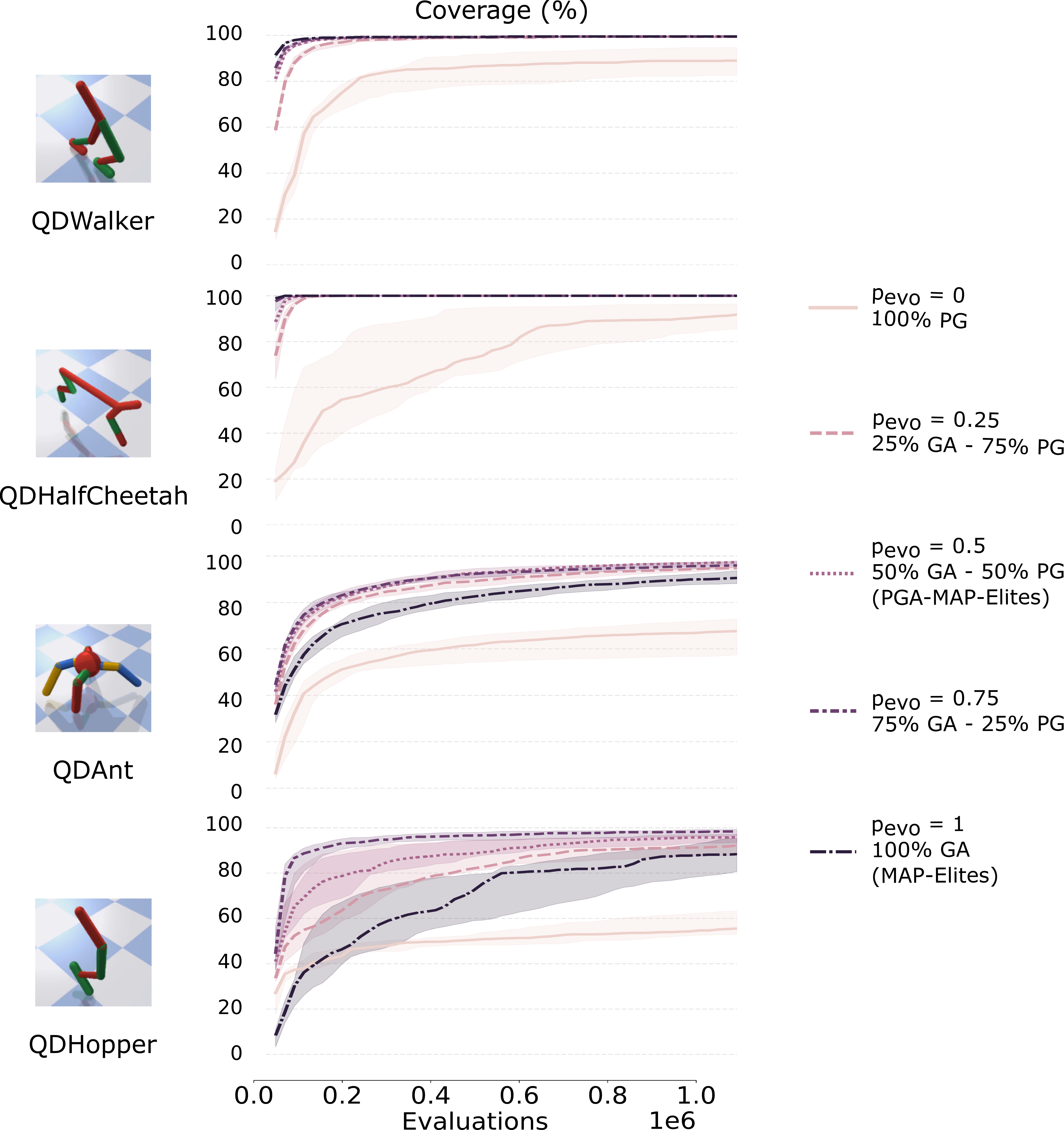}
\caption{
\textbf{Ablation $\nevo$ - Coverage:}
Coverage of \name{} with different mutation proportion on each QDGym task. 
Each experiment is replicated $10$ times, the solid line corresponds to the median over replications and the shaded area to the first and third quartiles.
}
\label{fig:progress_ablation_coverage}
\end{figure*}

\begin{figure*}[h!]
\centering
\includegraphics[width = 0.95\hsize]{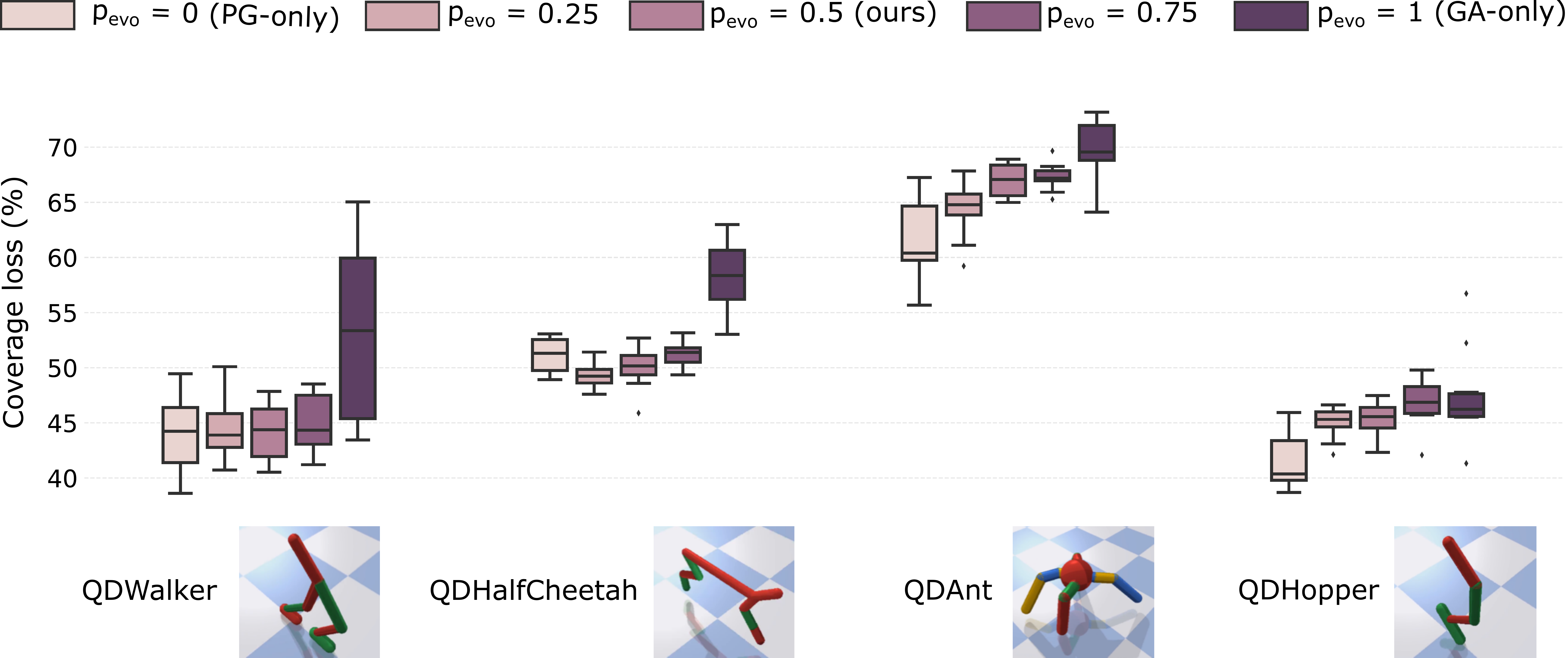}
\caption{
    \textbf{Ablation $\nevo$ - Loss comparison:}
    Comparison of the Coverage loss for \name{} with different variation operator proportions. 
    To compute these losses, all solutions of the archive are replaced in a Corrected archive according to their average BD and fitness over $50$ replications. The loss corresponds to the difference of each metric in proportion of the original metric.
    Lower loss indicates better reproducibility.
    The boxes represent the distribution of the loss for a given algorithm across $10$ runs.
}
\label{fig:loss_ablation_coverage}
\end{figure*}

We display in Fig. \ref{fig:progress_ablation_coverage} the Coverage metrics from Section \ref{sec:ablation} and in Fig. \ref{fig:loss_ablation_coverage} the corresponding Coverage loss.
These results lead to the same conclusion as the QD-Score and Max-Fitness analysis in Section \ref{sec:ablation}: the performance of \name{} are robust to the proportion of PG and GA, as long as the two operators are present.

\section{Wilcoxon rank-sum test p-values} \label{app:p-values}

In Table \ref{tab:p_values}, we give all the p-values based on Wilcoxon rank-sum test associated with all comparisons of \name{} with other baselines in each result Section.

\begin{table}[h]
  \caption{\textbf{Summary of p-values:} we give the p-value based on Wilcoxon rank-sum test associated with all comparisons of \name{} with other baselines in each result Section. ME stands for MAP-Elites.}
  
  \label{tab:p_values}
  
  \begin{tabular}{ c | c | c c c c c}
    \toprule
    \multicolumn{2}{c|}{Section \ref{sec:deterministic}} & ME-ES & ME & TD3 & QD-PG \\
    \midrule
    \multirow{2}{*}{QDWalker} & QD-Score & 1.38208e-14 & 1.38208e-14 & 1.38208e-14 & 3.55295e-10 \\
                           & Max-Fitness & 1.38208e-14 & 1.38208e-14 & 1.38208e-14 & 3.55295e-10 \\
    \multirow{2}{*}{QDHalfCheetah} & QD-Score & 3.01981e-14 & 1.38208e-14 & 1.38208e-14 & 1.86346e-07 \\
                           & Max-Fitness & 5.58359e-14 & 1.38208e-14 & 0.0153138 & 1.86346e-07 \\     
    \multirow{2}{*}{QDAnt} & QD-Score & 1.38208e-14 & 1.38208e-14 & 3.55295e-10 & 1.38208e-14 \\
                           & Max-Fitness & 3.47414e-13 & 1.38208e-14 & 0.0903488 & 3.55295e-10 \\      
    \multirow{2}{*}{QDHopper} & QD-Score & 3.01981e-14 & 1.38208e-14 & 1.38208e-14 & 3.55295e-10 \\
                           & Max-Fitness & 3.01981e-14 & 1.38208e-14 & 1.38208e-14 & 3.55295e-10 \\                     
    \midrule
    Deterministic & QD-Score & 1.38208e-14 & 1.38208e-14 & 3.55295e-10 & 3.01981e-14 \\
    QDWalker  & Max-Fitness & 1.38208e-14 & 1.38208e-14 & 3.42566e-08 & 3.55295e-10 \\
    Deterministic & QD-Score & 1.38208e-14 & 1.38208e-14 & 3.01981e-14 & 3.55295e-10 \\
    QDHalfCheetah  & Max-Fitness & 1.38208e-14 & 1.38208e-14 & 1.71916e-08 & 3.55295e-10 \\
    Deterministic & QD-Score & 1.38208e-14 & 3.01981e-14 & 1.38208e-14 & 3.55295e-10 \\
    QDAnt  & Max-Fitness & 1.38208e-14 & 3.01981e-14 & 0.0646717 & 3.55295e-10 \\
    Deterministic & QD-Score & 1.38208e-14 & 1.38208e-14 & 1.38208e-14 & 3.55295e-10 \\
    QDHopper  & Max-Fitness & 1.38208e-14 & 1.38208e-14 & 1.38208e-14 & 3.55295e-10 \\                       
    \bottomrule
    \end{tabular}
    
    \vspace{1em}

    \begin{tabular}{ c | c | c c c c}
    \toprule
    \multicolumn{2}{c|}{Section \ref{sec:stochastic}} & ME-ES & ME & Deep-grid & ME-sampling \\
    \midrule
    \multirow{2}{*}{QDWalker} & QD-Score & 1.38208e-14 & 1.38208e-14 & 3.55295e-10 & 3.55295e-10 \\
                           & Max-Fitness & 1.38208e-14 & 1.38208e-14 & 3.55295e-10 & 3.55295e-10 \\
    \multirow{2}{*}{QDHalfCheetah} & QD-Score & 3.01981e-14 & 1.38208e-14 & 3.55295e-10 & 3.55295e-10 \\
                           & Max-Fitness & 5.58359e-14 & 1.38208e-14 & 3.55295e-10 & 3.55295e-10 \\
    \multirow{2}{*}{QDAnt} & QD-Score & 1.38208e-14 & 1.38208e-14 & 3.55295e-10 & 3.55295e-10 \\
                           & Max-Fitness & 3.47414e-13 & 1.38208e-14 & 3.55295e-10 & 3.55295e-10 \\
    \multirow{2}{*}{QDHopper} & QD-Score & 3.01981e-14 & 1.38208e-14 & 3.55295e-10 & 3.55295e-10 \\
                           & Max-Fitness & 3.01981e-14 & 1.38208e-14 & 3.55295e-10 & 3.55295e-10 \\                
    \bottomrule
    \end{tabular}
    
    \vspace{1em}
    
    \begin{tabular}{ c | c | c c c c}
    \toprule
    \multicolumn{2}{c|}{Section \ref{sec:ablation}} & $\nevo{} = 0$ (Full PG) & $\nevo{} = 0.25$ & $\nevo{} = 0.75$ & $\nevo{} = 1$ (Full GA) \\
    \midrule
    \multirow{2}{*}{QDWalker} & QD-Score & 0.000379285 & 0.000880743 & 0.173617 & 0.000157052 \\
                           & Max-Fitness & 0.000529757 & 0.0126111 & 0.364346 & 0.000157052 \\
    \multirow{2}{*}{QDHalfCheetah} & QD-Score & 0.000157052 & 0.0864107 & 0.705457 & 0.000238563 \\
                           & Max-Fitness & 0.000157052 & 0.567628 & 0.198765 & 0.000238563 \\
    \multirow{2}{*}{QDAnt} & QD-Score & 0.000157052 & 0.596701 & 0.00515896 & 0.000157052 \\
                           & Max-Fitness & 0.762369 & 0.0696424 & 0.0126111 & 0.000157052 \\
    \multirow{2}{*}{QDHopper} & QD-Score & 0.000157052 & 0.0342937 & 0.0101652 & 0.000157052 \\
                           & Max-Fitness & 0.00149887 & 0.449692 & 0.04125 & 0.000157052 \\                
    \bottomrule
    \end{tabular}

\end{table}


\end{document}